\documentclass[letterpaper]{article}
\pdfoutput=1
\usepackage{aaai24}  % DO NOT CHANGE THIS
\usepackage{times}  % DO NOT CHANGE THIS
\usepackage{helvet}  % DO NOT CHANGE THIS
\usepackage{courier}  % DO NOT CHANGE THIS
\usepackage[hyphens]{url}  % DO NOT CHANGE THIS
\usepackage{graphicx} % DO NOT CHANGE THIS
\usepackage{natbib}  % DO NOT CHANGE THIS AND DO NOT ADD ANY OPTIONS TO IT
\usepackage{caption} % DO NOT CHANGE THIS AND DO NOT ADD ANY OPTIONS TO IT
\usepackage{algorithm}
\usepackage{algorithmic}
\usepackage{bbding}
\usepackage{newfloat}
\usepackage{listings}
\DeclareCaptionStyle{ruled}{labelfont=normalfont,labelsep=colon,strut=off} % DO NOT CHANGE THIS
\floatstyle{ruled}
\newfloat{listing}{tb}{lst}{}
\floatname{listing}{Listing}
\title{Decoupled Textual Embeddings for Customized Image Generation}
\author{
    Yufei Cai\textsuperscript{\rm 1,\rm 2},
    Yuxiang Wei\textsuperscript{\rm 2}, Zhilong Ji\textsuperscript{\rm 3}, Jinfeng Bai\textsuperscript{\rm 3}, 
    Hu Han\textsuperscript{\rm 1}, Wangmeng Zuo\textsuperscript{\rm 2,\rm 4}
}
\affiliations{
    \textsuperscript{\rm 1}Institute of Computing Technology, Chinese Academy of Sciences\\
    \textsuperscript{\rm 2}Harbin Institute of Technology, 
    \textsuperscript{\rm 3}Tomorrow Advancing Life, \textsuperscript{\rm 4}Pengcheng Lab \\

    caiyufei23@mails.ucas.ac.cn, yuxiang.wei.cs@gmail.com, zhilongji@hotmail.com, jfbai.bit@gmail.com \\ hanhu@ict.ac.cn, wmzuo@hit.edu.cn
}

\usepackage{amssymb}
\usepackage{booktabs}   
\usepackage{amsmath}
\usepackage{makecell}

\usepackage{color}

\makeatletter
\DeclareRobustCommand\onedot{\futurelet\@let@token\@onedot}
\def\@onedot{\ifx\@let@token.\else.\null\fi\xspace}

\makeatother
\begin{document}
\maketitle

\begin{figure*}[t]
    \centering
    \includegraphics[width=1\linewidth]{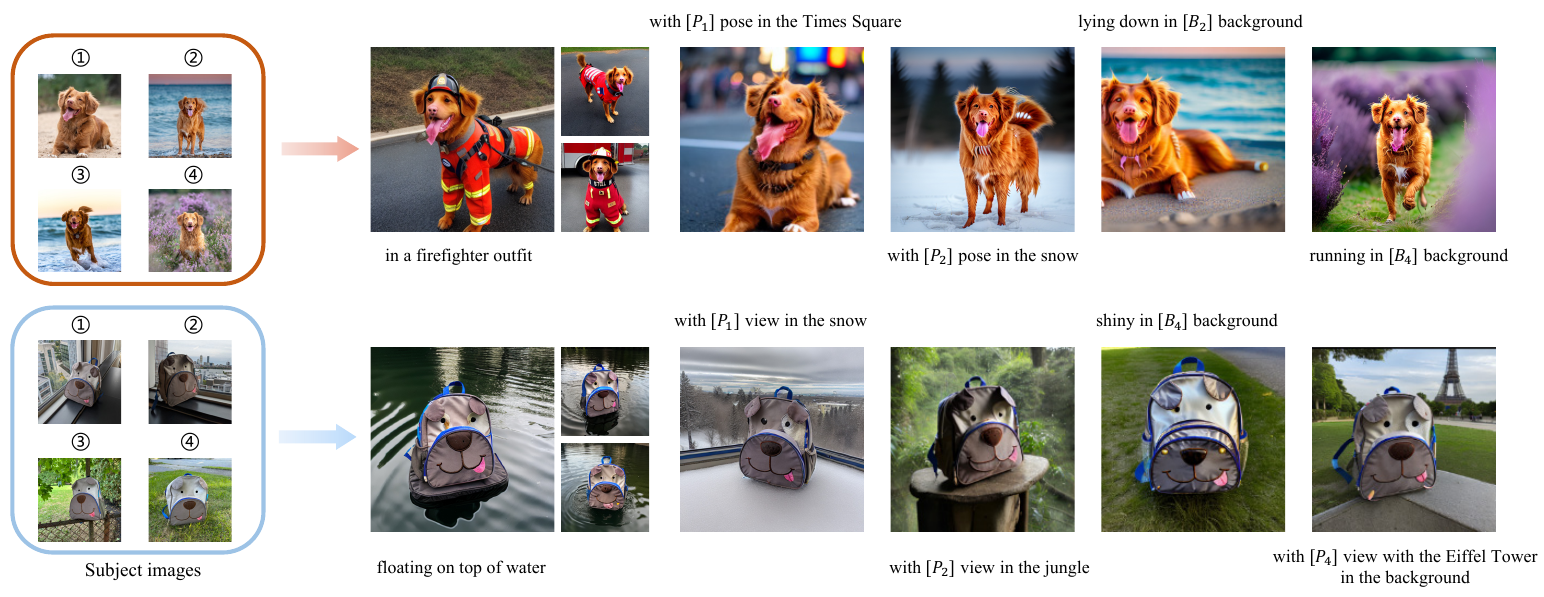}
    \vspace{-2em}
    \caption{\textbf{Customized images generated by our DETEX.} The identifier \texttt{[$B_i$]} (\texttt{[$P_i$]}) indicates the learned background (pose) of the $i$-th input image. Our DETEX can generate high-quality and diverse images when composing the learned concept into a new scene, while allowing users to selectively retain specific attribute information for controllable generation.}
%    \vspace{-1.5em}
    \label{fig:intro}
\end{figure*}

\begin{abstract}

Customized text-to-image generation, which aims to learn user-specified concepts with a few images, has drawn significant attention recently.
However, existing methods usually suffer from overfitting issues and entangle the subject-unrelated information (\textit{e.g}, background and pose) with the learned concept, limiting the potential to compose concept into new scenes.
To address these issues, we propose the DETEX, a novel approach that learns the disentangled concept embedding for flexible customized text-to-image generation.
Unlike conventional methods that learn a single concept embedding from the given images, our DETEX represents each image using multiple word embeddings during training, \textit{i.e}, a learnable image-shared subject embedding and several image-specific subject-unrelated embeddings.
To decouple irrelevant attributes (\textit{i.e}, background and pose) from the subject embedding, we further present several attribute mappers that encode each image as several image-specific subject-unrelated embeddings.
To encourage these unrelated embeddings to capture the irrelevant information, we incorporate them with corresponding attribute words and propose a joint training strategy to facilitate the disentanglement.
During inference, we only use the subject embedding for image generation, while selectively using image-specific embeddings to retain image-specified attributes. 
Extensive experiments demonstrate that the subject embedding obtained by our method can faithfully represent the target concept, while showing superior editability compared to the state-of-the-art methods.
Our code will be available at \url{https://github.com/PrototypeNx/DETEX}.

\end{abstract}

\section{1 Introduction}

Recently, diffusion models~\cite{DALLE2,Imagen,LDM,GLIDE} have demonstrated remarkable superiority in text-to-image generation.
Benefiting from the large-scale pretraining, these models can generate diverse and photorealistic images based on textual descriptions, showing great potential in various tasks, such as image manipulation and artistic creation.

Besides text-to-image generation, substantial efforts have been devoted to customized image generation~\cite{TI}, which aims to learn user-specified concept from a small set of images describing target concept (typically 3-5 images).
Existing methods for customized text-to-image generation~\cite{CD,dreambooth,TI} usually aligned the target concept with a user-specified word by finetuning the word embedding or model parameters.
However, due to the limited training examples, the learned concept inevitably contains subject-unrelated information (\textit{e.g}, image background, subject pose and position), resulting in degenerated editability.
Although some studies~\cite{avrahami2023break, ELITE} employed a subject mask to filter the background information, the gain was unsatisfactory, as the irrelevant disturbances (\textit{e.g}, blank background and pose) still entangled with the learned concept.
Disenbooth~\cite{disenbooth} decomposed the target concept into a subject embedding and an additional irrelevant embedding to exclude the irrelevant information. 
However, there lacks explicit supervision to effectively facilitate the decoupling between the subject concept and the irrelevant information.
% Disenbooth~\cite{disenbooth} explored the decomposition of text embeddings into two components: relevant and irrelevant, aiming to mitigate the interference of irrelevant information. However, Disenbooth lacks explicit supervision at the text embedding level, which prevents it from achieving a finer decoupling between relevant and irrelevant information.

To address the above issues, we propose DETEX, a method that learns disentangled concept embedding for flexible customized text-to-image generation.
Following the principles of Custom Diffusion~\cite{CD}, we adapt the pretrained Stable Diffusion~\cite{LDM} model to new concept by finetuning both model parameters and the word embedding.
Instead of learning a single concept embedding to represent all given images, our DETEX represents the target concept and subject-unrelated information using separate word embeddings, \textit{i.e}, an image-shared subject embedding and several image-specific subject-unrelated embeddings.
Specifically, we introduce a learnable image-shared subject embedding to represent the target concept.
To decouple irrelevant attributes from the subject embedding, for each input image, we additionally introduce several image-specific subject-unrelated embeddings to capture the irrelevant information.
Here, we consider irrelevant information from two main aspects: pose (view) and background.
The corresponding attribute words (\textit{i.e}, \texttt{[B] background} and \texttt{[P] pose/view}) are incorporated with the embeddings to facilitate them capturing the irrelevant information.
% 
% Overall, we aim to have the learned shared token solely describe the target concept without carrying overfitted irrelevant information. 

To effectively decouple the target concept with unrelated pose and background information, we propose a joint training strategy.
Specifically, we encourage different embedding to reconstruct the corresponding information representation based on the given images and their subject masks.
However, directly optimizing the image-specific subject-unrelated embeddings is insufficient to capture specific attributes efficiently. 
To address this limitation, we propose an attribute mapper for each attribute which projects each image as the corresponding subject-unrelated embedding. 
The CLIP~\cite{CLIP} image encoder serves as the feature extractor in this process.
During inference, only the subject embedding is utilized for image generation.
Our experimental results show that our learned subject embedding can faithfully represent the target concept, and possess excellent editability compared to the state-of-the-art (SOTA) methods. 
Furthermore, we can selectively retain specific irrelevant attribute information by keeping corresponding unrelated embeddings, allowing users to flexibly control the image generation (see Fig.~\ref{fig:intro}).

Our contributions can be summarized as follows:
\begin{itemize}
\setlength{\itemsep}{1pt}
\setlength{\parsep}{0pt}
\setlength{\parskip}{0pt}
\item We propose a customized image generation method that utilizes multiple tokens to alleviate the issue of overfitting and entanglement between the target concept and unrelated information.
\item Our method enables more precise and efficient control over preserving input image content in the generated results during inference by selectively utilizing different tokens. This enhances the controllability of personalized generation.
\item Extensive experiments show that our method outperforms the SOTA methods in terms of editing flexibility. Furthermore, our method exhibits stronger editing potential, especially when the number of input images is extremely limited.
\end{itemize}

In conclusion, this work proposes a novel solution to customized generation in text-to-image models, addressing the challenges of overfitting and controllability.

\begin{figure*}[t]
    \centering
    \includegraphics[width=1\linewidth]{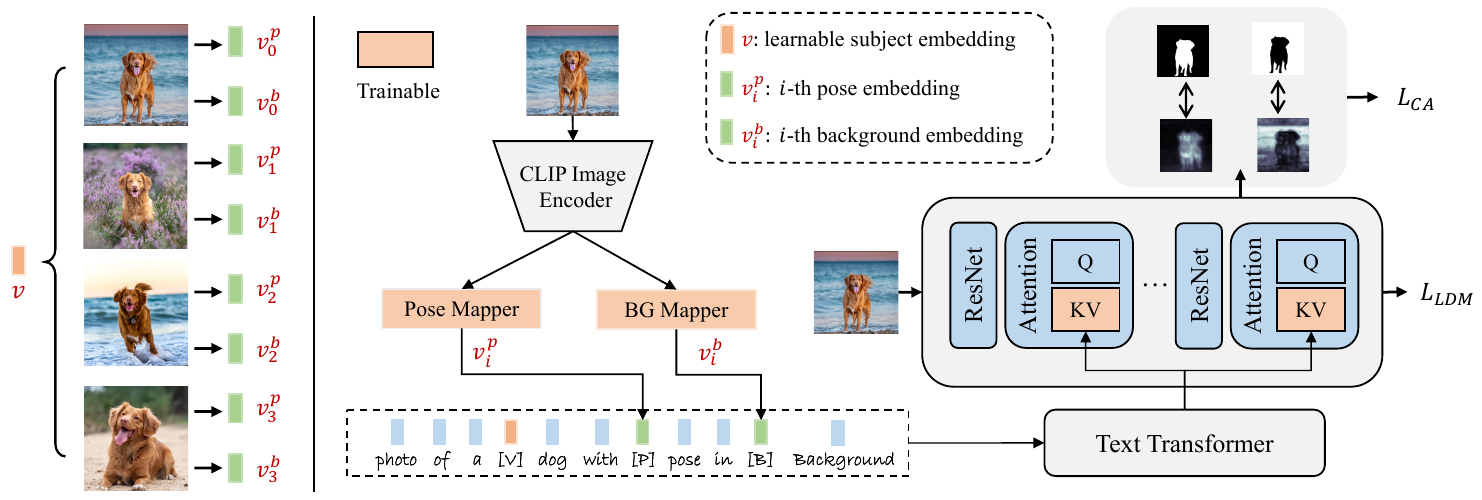}
    \vspace{-1.8em}
    \caption{\textbf{Framework of our DETEX.} \textbf{Left}: Our DETEX represents each image with multiple decoupled textual embeddings, \textit{i.e}, an image-shared subject embedding $v$ and two image-specific subject-unrelated embeddings (pose $v^p_i$ and background $v^b_i$). \textbf{Right}: To learn target concept, we initialize the subject embedding $v$ as a learnable vector, and adopt two attribute mappers to project the input image as the pose and background embeddings. During training, we jointly finetune the embeddings with the K, V mapping parameters in cross-attention layer. A cross-attention loss is further introduced to facilitate the disentanglement.}
    \vspace{-1.2em}
    \label{fig_pipeline}
\end{figure*}

\section{2 Related Work}

\noindent \textbf{Text-to-Image Diffusion Models.}
Diffusion models \cite{DDPM,diffusion} are parametric neural networks that learn the data distribution by progressive denoising. It has achieved remarkable performance in the field of image synthesis. Subsequent works have focused on exploring diffusion-based conditional generation. Classifier-free guidance \cite{classifier-free} based on the derivation of conditional probabilities is a direct conditional generation approach that jointly optimizes over a large amount of data, resulting in reliable and detail-preserving generation effects. Benefiting from the advancements in language models and multi-modal models such as CLIP \cite{CLIP} and BERT \cite{BERT}, further work has been devoted to diffusion-based text-to-image generation. Some work \cite{blended—diffusion,DiffusionCLIP} use text as a condition to control image synthesis. By leveraging the powerful information extraction capability of language models and the classifier-free guidance techniques, current approaches \cite{DALLE2,Imagen,GLIDE} promote semantic alignment between cross-modal information. Through large-scale training on extensive data samples, they have achieved impressive text-to-image synthesis results. Latent diffusion models \cite{LDM} compress data into a low-dimensional latent space, finding an optimal balance between computational complexity reduction and detail preservation. Text-to-image models based on LDM, such as Stable Diffusion, exhibit more efficient text-driven synthesis capabilities, reaching high levels of diversity and generality.

Despite the impressive performance of existing large-scale text-to-image models, they solely rely on natural language prompts guidance and cannot perform customized image generation. Specifically, it is extremely challenging to consistently maintain the identity of a specific concept in the images generated by such models. This has led to the emergence of customized image generation tasks.

\noindent \textbf{Customized Image Generation.}
Current customized generation methods primarily rely on fine-tuning. Textual Inversion \cite{TI} only fine-tunes word embeddings to learn the target concepts. However, it lacks the ability to fit new concepts effectively and struggles to capture fine-grained details of concepts in various conditions. DreamBooth \cite{dreambooth} fine-tunes the full weights of the U-net and text encoder to achieve more effective concept-fitting ability, but it introduces language drifts and information forgetting, leading to poorer editing flexibility and controllability. Subsequent work attempts to find a more compact and efficient parameter space for fine-tuning to alleviate model overfitting. Custom Diffusion \cite{CD} only fine-tunes the weights of the cross-attention layers, while SVDiff \cite{svdiff} fine-tunes the spectral space of weight matrices. Later, non-fine-tuning methods for customized generation emerged. Cones \cite{cones} focuses on identifying the effective concept neurons related to the target concept, while ViCo \cite{vico} proposes a plug-in image attention module to adjust the diffusion process. Other works \cite{ELITE, Instantbooth} explore achieving customized generation without finetuning. These encoder-based approaches propose an additional module pretrained on additional datasets to reduce the time cost of generation significantly. DisenBooth \cite{disenbooth} explores the decoupling of identity-irrelevant information from the target concept during customized fintune. It decomposes text embeddings into identity-related and identity-irrelevant parts to generate new images. In this paper, we model the pose and background information as independent image-specific words, achieving finer decoupling at the word embedding stage. Our approach enables more flexible editing ability and precise control over retaining inherent content from input images during inference.

\vspace{-0.5em}
\section{3 Proposed Method}

\subsection{3.1 Preliminary}

In this work, we employ the pretrained Stable Diffusion (SD)~\cite{LDM} as our text-to-image model, and adapt it for customized text-to-image generation.
In the following, we will give a brief introduction of the SD and our baseline method for customized text-to-image generation, \textit{i.e}, Custom Diffusion~\cite{CD}.

\noindent \textbf{Stable Diffusion.}  Stable Diffusion~\cite{LDM} is trained on large-scale data and comprises two components.
First, an autoencoder ($\mathcal{E}(\cdot)$, $\mathcal{D}(\cdot)$) is trained to map an image $x$ to a lower dimensional latent space by the encoder $z = \mathcal{E}(x)$, and then reconstructed back to the image by the decoder $ D(\mathcal{E}(x)) \approx x $.
Then, the conditional diffusion model $\epsilon_{\theta}(\cdot)$ is trained 
on the latent space to generate latent codes based on text condition $y$.
To train the diffusion model, a simple mean-squared loss is adopted,
\begin{equation}
\small
L_{LDM} = \mathbb{E}_{z\sim\mathcal{E}(x), y, \epsilon \sim \mathcal{N}(0, 1), t }\Big[ \Vert \epsilon - \epsilon_\theta(z_{t},t, \tau_\theta(y)) \Vert_{2}^{2}\Big] \, ,
\label{eq:LDM_loss}
\end{equation}
where $\epsilon$ denotes unscaled noise, $t$ is time step, $z_t$ is latent noised to time $t$, $\tau_\theta(\cdot)$ represents the pretrained CLIP text encoder~\cite{CLIP}.
During inference, a random Gaussian noise $z_T$ is iteratively denoised to $z_0$, ultimately yielding the final image through the decoder $x' = \mathcal{D}(z_0)$.

\noindent \textbf{Customized Text-to-Image Generation.} 
Given a small set of images $\{ x_i \}_{i=1}^N$ describing the user-specified concept, 
Custom Diffusion~\cite{CD} first introduces a learnable word embedding $v$ to represent the target concept.
A pseudo-word \texttt{[V]} is further introduced to the vocabulary and $v$ is associated as its word embedding.
With \texttt{[V]}, we can generate new images with desired concept, such as ``\texttt{A [V] swimming}''.
During training, Custom Diffusion simultaneously optimizes $v$ and the parameters of K, V mapping in cross-attention layers by minimizing Eqn.~\ref{eq:LDM_loss} over the given images.
To incorporate $v$ into the generation, the input text prompt $y$ is formulated as ``\texttt{Photo of a [V] [class]}'', where \texttt{[class]} is the category prior to the given subject.
Furthermore, a prior preserving loss $L_{pr}$ is adopted by Custom Diffusion to preserve the prior characteristics of pretrained SD model,
\begin{equation}
\small
L_{pr} = \mathbb{E}_{z\sim\mathcal{E}(x_{pr}), y_{pr}, \epsilon \sim \mathcal{N}(0, 1), t }\Big[ \Vert \epsilon - \epsilon_\theta(z_{t},t, \tau_\theta(y_{pr})) \Vert_{2}^{2}\Big] \, ,
\label{eq:PR_loss}
\end{equation}
The regularization images $\{x_{pr}\}$ are retrieved from LAION-400M \cite{LAION-400M} dataset or generated by original SD model, while the regularization prompt $y_{pr}$ is set as ``\texttt{Photo of a [class]}''.

\subsection{3.2 Decoupled Textual Embeddings}

Custom Diffusion learns a novel subject embedding to represent all images.
However, due to the limited training examples, the learned subject embedding inevitably contains subject-unrelated information (\textit{e.g}, image background, subject pose, and position), thereby limiting its ability in composing novel scenes.
To address this issue, we propose a novel approach DETEX, which learns the decoupled textual embedding for flexible customized text-to-image generation.
Following the principles of Custom Diffusion~\cite{CD}, we adapt the pretrained Stable Diffusion~\cite{LDM} model to learn new concept by finetuning both model parameters and the image-shared subject embedding $v$.
To decouple irrelevant information from the subject embedding, we introduce multiple image-specific subject-unrelated embeddings.
Here, we consider two primary sources of irrelevant information: pose (view) and background.
For each input image $x_i$, we introduce two image-specific subject-unrelated embeddings $v^p_i$ (\texttt{[$P_i$]}) and $v^b_i$ (\texttt{[$B_i$]}) designated to represent the pose and background information, respectively.
\texttt{[$P_i$]} and \texttt{[$B_i$]} are the corresponding pseudo-words.
These embeddings are incorporated with specified attribute words to capture the corresponding information, which takes the form of ``\texttt{Photo of a [V] class with [P] pose/view in [B] background}''.

Since the pose and background may vary significantly across different images, directly optimizing the corresponding subject-unrelated embeddings is usually non-trivial. 
%
% insufficient to capture specific attributes efficiently
%
To facilitate the embedding learning,  we propose an embedding mapper for each attribute that projects each image as the corresponding subject-unrelated embedding, as shown in Fig.~\ref{fig_pipeline}. 
Specifically, we leverage a pretrained CLIP~\cite{CLIP} image encoder to extract features for image $x_i$.
Then, the pose mapper projects the CLIP feature to subject-unrelated pose embedding, 
\begin{equation}
v^p_i= M^P_i (E_I (x_i)),
\end{equation}
where $E_I(\cdot)$ is the pretrained CLIP image encoder. 
$M^P_i$ denotes the pose mapper, which is implemented as a small Multilayer Perceptron (MLP). 
Similarly, we can obtain the subject-unrelated background embedding $v^b_i$ through the background mapper $M^B_i$.

\subsection{3.3 Joint Training Strategy}

To effectively decouple the subject concept and unrelated information, we further propose a cross-attention loss alongside a joint training strategy.
Intuitively, if the pose and background embeddings are well disentangled, they should focus on the respective image regions during image generation.
Therefore, we first introduce a cross-attention loss that encourages different embedding to reconstruct the related information. 
In particular, we employ a subject mask that highlights the target concept and the background region.
%
% , and apply a constraint on the cross-attention map.
%
During training, we constrain the cross attention map of $[P]$ word to align with the subject region, while the attention map of $[B]$ word aligns with the background region,
\begin{equation}
L_{CA}=\mathbb{E}_{i,z,t}[||A(P_i,z_t) 
-m_i||+||A(B_i,z_t) - \overline{m_i}||], \label{eqn:CA}
\end{equation}
where $A(P_i,z_t)$ and $A(B_i,z_t)$ are the cross attention maps \textit{w.r.t} , $B_i$ and $P_i$, respectively. 
$m_i$ denotes the subject mask corresponding to input image $x_i$ and $\overline{m_i} = 1 - m_i$. 
In conclusion, our overall training objective is,
\begin{equation}
L = L_{LDM} + \lambda_{pr} L_{pr} + \lambda_{CA} L_{CA},
\end{equation}
where $\lambda_{pr}$ and $\lambda_{pr}$ are weights balancing different losses. 
In our implementation, we set $\lambda_{pr} = 1.0$ and $\lambda_{CA} = 0.01$.

Additionally, we present a joint training strategy to further enhance the disentanglement.
Specifically, we introduce a random image background filtering mechanism with a probability of $\gamma$ during training to prevent the subject embedding and pose embedding from entangling the background information.
In this case, we set the input text prompt $y'$ as ``\texttt{Photo of a [V] class with [P] pose/view}'', while using the background-masked image $\{x_i^{'}\}_{i=1}^N$ as model input.
To achieve this, we pre-process the input image set $\{x_i\}_{i=1}^N$ using the subject mask $\{m_i\}_{i=1}^N$, resulting in foreground subject images with a blank background.
With this, our method can effectively disentangle the subject-unrelated pose and background information, obtaining a well-editable subject concept for flexible customized text-to-image generation.

\begin{figure*}[t]
    \centering
    \includegraphics[width=1\linewidth]{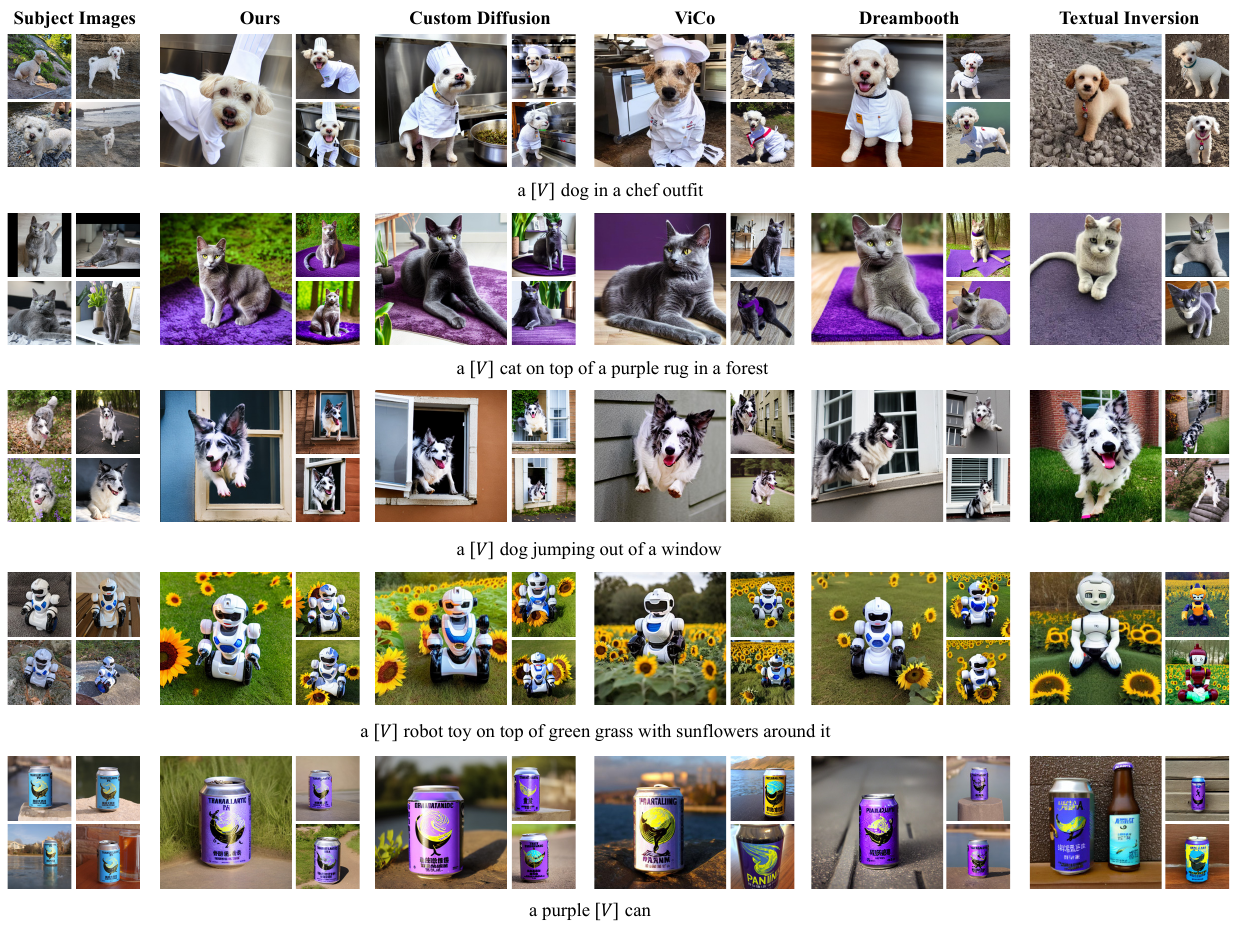}
    \vspace{-2em}
    \caption{\textbf{Visual comparisons.} Our DETEX demonstrates superior concept editability and subject fidelity compared to Textual Inversion~\cite{TI}, Dreambooth~\cite{dreambooth}, Custom Diffusion~\cite{CD}, and ViCo~\cite{vico}.}
    \vspace{-1em}
    \label{fig_qualitive}
\end{figure*}

\section{4 Experiments}

\subsection{4.1 Experimental Details}

\noindent \textbf{Datasets.}
We conduct our experiment on the DreamBench \cite{dreambooth} dataset. 
It consists of 30 subjects with different categories (\textit{e.g}, animals, toys, and wearable items), and each subject has 4 $\sim$ 7 images.
We randomly filter out some images to ensure that each subject only keeps 4 images for training.
There are 25 editing prompts for each subject. 
For evaluation, we randomly generate 8 images for each subject-prompt pair, obtaining 6,000 images in total.

\noindent \textbf{Metrics.}
Following Dreambooth~\cite{dreambooth}, we evaluate our method with three metrics: CLIP-T, CLIP-I and DINO-I.
CLIP-T calculates the feature similarity between the CLIP visual feature of the generated image and the CLIP textual feature of the corresponding prompt text.
We omit the placeholder and keep the class name when calculating the textual feature.
CLIP-I calculates the CLIP visual similarity between the generated and target concept image.
DINO-I calculates the feature similarity between the ViTS/16 DINO~\cite{DINO} embeddings of generated and concept images.

\noindent \textbf{Implementation Details.}
We employ Stable Diffusion v1-4 as our pretrained text-to-image model.
The training process is conducted on RTX 3090 using AdamW ~\cite{adamw} optimizer with a batch size of 4 for 600 steps.
The learning rate is set as 1e-5, and the drop probability $\gamma$ is set as 0.5.
Our embedding mappers are implemented as the 3-layer MLPs, and each has a size of 6.7 MB.
The cross attention loss (Eqn.~\ref{eqn:CA}) is calculated at the resolution of 32 $\times$ 32, and we average the attention maps along the head dimension.
During testing, we generate images with 50 DDIM~\cite{DDIM} steps, and the scale of classifier-free guidance is 6. 

\subsection{4.2 Qualitative Evaluation}
We first qualitatively compare our DETEX with existing methods, including Textual Inversion~\cite{TI}, Dreambooth~\cite{dreambooth}, Custom Diffusion~\cite{CD}, SVDiff~\cite{svdiff}, and ViCo~\cite{vico}.
The comparisons are illustrated in Fig.~\ref{fig_qualitive}.
We can see, our method shows superior ability in generating images with higher subject fidelity and text alignment.
For subject fidelity, our subject embedding $v$ faithfully captures the target concept, and generates image with concise details (\textit{e.g}, the appearance of the robot on the $4th$ row and the pattern on the bottle body on the $5th$ row).
For text alignment, our method exhibits flexible editability, allowing it to compose into new scene.
For instance, with the prompt ``\texttt{on top of a purple rug in a forest}'' ($2nd$ row), our method can generate images that align well with text, whereas the competitors may overlook the ``\texttt{in a forest}''.
Furthermore, we can selectively retain specific irrelevant attributes by keeping corresponding unrelated embeddings, thereby allowing users to flexibly control the image generation (see Fig.~\ref{fig:intro}).
More qualitative results are provided in the supplementary materials.

\begin{table*}[t]
    % \captionsetup{font={small, stretch=1.1}, justification=raggedright}
    \centering
    \caption{\textbf{Quantitative comparisons with existing methods.} For subject fidelity metrics, we compute on both the original input images (CLIP-I and DINO-I) and foreground-only input images (CLIP-I (FG) and DINO-I (FG)).}
    \setlength{\tabcolsep}{3mm}
    \label{tab:quantitative}  
    \resizebox{1\linewidth}{!}{
        \begin{tabular}{cccccc}
            \toprule
            Method & CLIP-T $(\uparrow)$ & CLIP-I $(\uparrow)$ & CLIP-I (FG) $(\uparrow)$ & DINO-I $(\uparrow)$ & DINO-I (FG) $(\uparrow)$ \\
            \midrule
            Textual Inversion~\cite{TI} & 0.2877& 0.6833 & 0.6793  & 0.5203 & 0.5178 \\ 
            Custom Diffusion~\cite{CD} & 0.3143 & 0.7870 & 0.7553 & 0.6602 & 0.5970 \\ 
            SVDiff~\cite{svdiff} & 0.3167 & 0.7782 & 0.7449 & 0.6297 & 0.5672 \\
            ViCo~\cite{vico} & 0.2908 & 0.7774 & 0.7445   & 0.6337 & 0.5642  \\
            Dreambooth~\cite{dreambooth} & 0.3113 & \textbf{0.7897} & 0.7499  & \textbf{0.6636} & 0.5981 \\
            Dreambooth+LoRA~\cite{lora} & 0.3123 & 0.7872 & 0.7498  & 0.6498 & 0.5710 \\
            Ours & \textbf{0.3249} & 0.7824 & \textbf{0.7688}  & 0.6504 & \textbf{0.6121} \\
            \bottomrule
        \end{tabular}
    }
\end{table*}

\begin{table}[t]
    \small
    \centering
    \caption{\textbf{User study.} The numbers indicate the percentage (\%) of volunteers who favor the results of our method over those of the competing methods based on the given question.}
    \setlength{\tabcolsep}{1mm}
    \label{tab5:user_study}  
    \resizebox{\linewidth}{!}{
    \begin{tabular}{ccccc}
        \toprule
        \thead{Method} & \thead{Ours vs.\\Custom Diffusion} & \thead{Ours vs.\\Dreambooth} & \thead{Ours vs.\\ViCo} & \thead{Ours vs.\\SVDiff}\\
        \midrule
        Text-alignment & 58.30 & 60.65 & 71.45 & 54.95 \\ 
        Image-alignment & 67.90 & 54.15 & 79.55 & 75.25 \\
        \bottomrule
    \end{tabular}
    }
\end{table}
\begin{figure*}[t]
    \centering
    \captionsetup{font={footnotesize}, justification=raggedright}
    \includegraphics[width=1\linewidth]{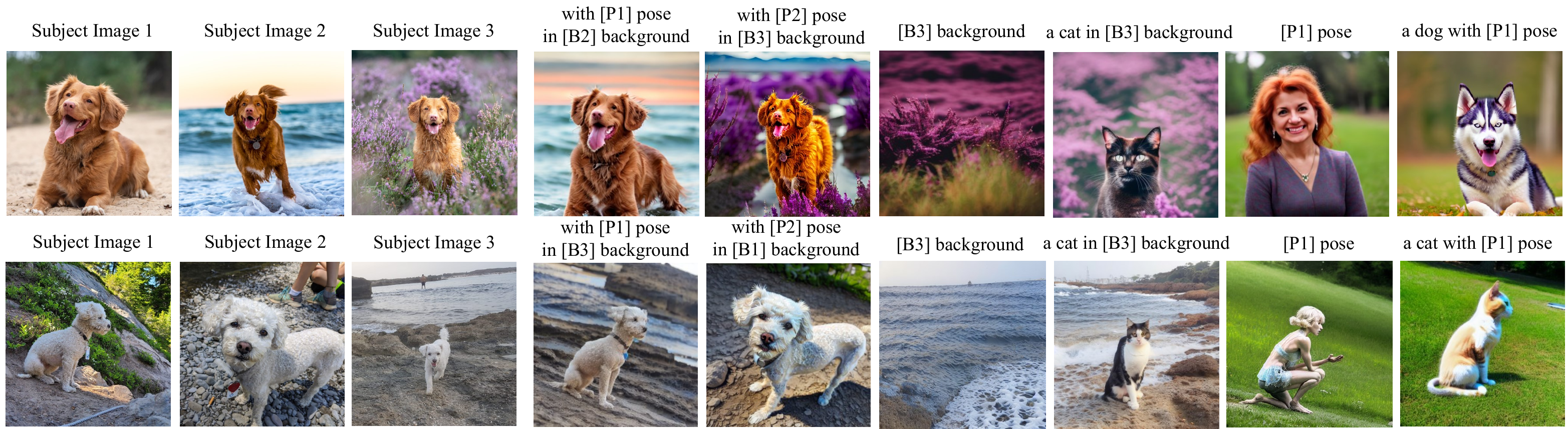}
    \vspace{-1.5em}
    \caption{\textbf{Visualization of decoupled textual embeddings}. The unrelated tokens [P] and [B] can be used with various combinations. The specific attributes are well maintained in the corresponding generation results.}
    \vspace{-1em}
    \label{fig:all}
\end{figure*}

\begin{figure*}[t]
    \centering
    \includegraphics[width=1\linewidth]{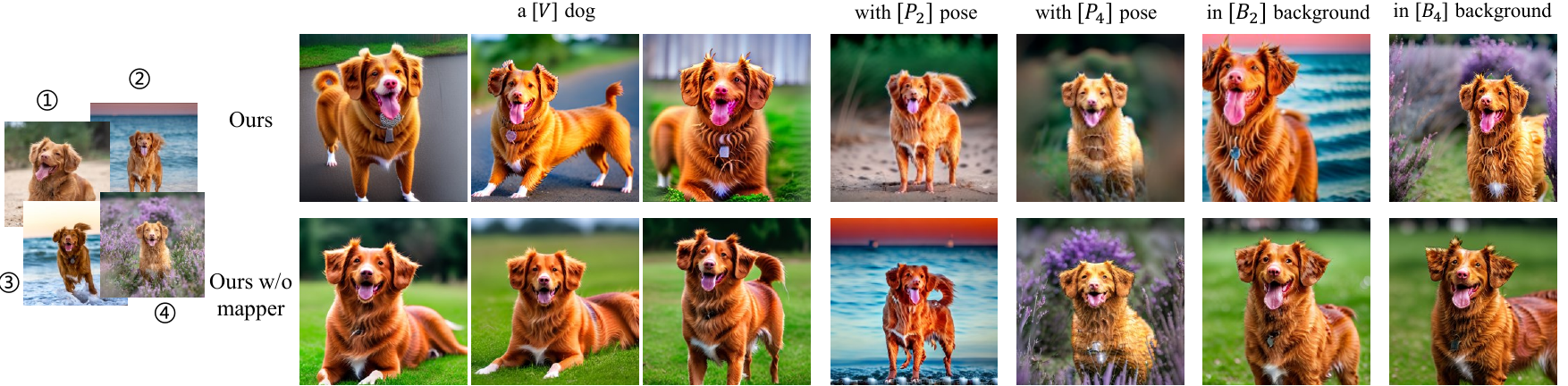}
    \vspace{-1.5em}
    \caption{\textbf{Visual comparisons on the effect of attribute mappers.} With the attribute mappers, our method can effectively disentangle the subject, pose, and background information. Our learned subject embedding can be used to generate target concept with various pose and background. The pose and background embeddings independently control the specific attributes.}
    \vspace{-1.5em}
    \label{fig:ablation_mapper}
\end{figure*}

\subsection{4.3 Quantitative Evaluation}

We further conduct the quantitative evaluation to validate the effectiveness of our DETEX.
As shown in Table~\ref{tab:quantitative}, our methods achieves
better text alignment (\textit{i.e}, CLIP-T) compared to the state-of-the-art methods, demonstrating its superior editability.
Although existing methods like Dreambooth and Custom Diffusion exhibit higher image alignment scores (\textit{i.e}, CLIP-I and DINO-I), these methods tend to entangle the subject-irrelevant information (\textit{e.g}, background) with the learned concept and inevitably contain these information in the generated images (\textit{e.g}, the plant of Custom Diffusion on the $2nd$ row).
Consequently, a direct comparison of CLIP-I and DINO-I metrics may be unfair since they are calculated on full images.
To alleviate this issue, we additionally calculate CLIP-I and DINO-I metrics solely within the foreground-subject region, denoted as CLIP-I(FG) and DINO-I(FG), respectively.
As shown in Table~\ref{tab:quantitative}, our method has better CLIP-I(FG) and DINO-I(FG), highlighting the ability to faithfully capture the target concept.
Furthermore, it is worth noting that the image alignment of existing methods consistently decreases after filtering out the background. 
In contrast, our method shows less decrease in values, thereby supporting our earlier analysis.

\noindent \textbf{User Study.}
We performed a user study to compare our approach with existing methods.
Given a subject, users were presented with two synthesized images, asked to select the better one from two aspects:
i) Text alignment: ``Which image is more consistent with the text?'';
ii) Image alignment: ``Which image better represents the objects in target images?''.
For each evaluated view, we employ 20 users, each user is asked to answer 400 randomly selected questions, resulting in 8000 responses in total.
As shown in Table~\ref{tab5:user_study}, our method receives more preference than other methods.

\subsection{4.4 Ablation Study}

We have conducted the ablation studies to evaluate the effects of various components in our method, including the multiple textual word embeddings, attribute mappers, the cross attention loss, and the joint training strategy.

\noindent \textbf{Effect of Decoupled Textual Embeddings.}
We first visualize the textual word embeddings learned by our method, and the results are illustrated in Fig.~\ref{fig:all}.
Our DETEX can combine the pose and background from different images with the target concept. 
Meanwhile, the unrelated tokens can be separately used to represent the specific attributes. 
They can also be combined with non-target concept.
All the results above demonstrate our method decouples the irrelevant information from the subject embedding successfully.
% This demonstrates our method decouples the irrelevant information from the subject embedding successfully.
% From the figure, when we generate images with only subject embedding (\textit{i.e}, \texttt{A [V] class}), the results contain the target subject with random pose and background.
% %
% This demonstrates our method decouples the irrelevant information from the subject embedding successfully.
% %
% Besides, we further introduce the learned pose embedding (\textit{i.e}, \texttt{A [V] class with [P] pose/view}), and found that our method can faithfully reconstruct the target pose without bringing the background information.
% %
% The background embedding is the same, demonstrating the effectiveness of our method to learn the decoupled textual word embeddings.

\begin{figure}[t]
    \centering
    \includegraphics[width=1\linewidth]{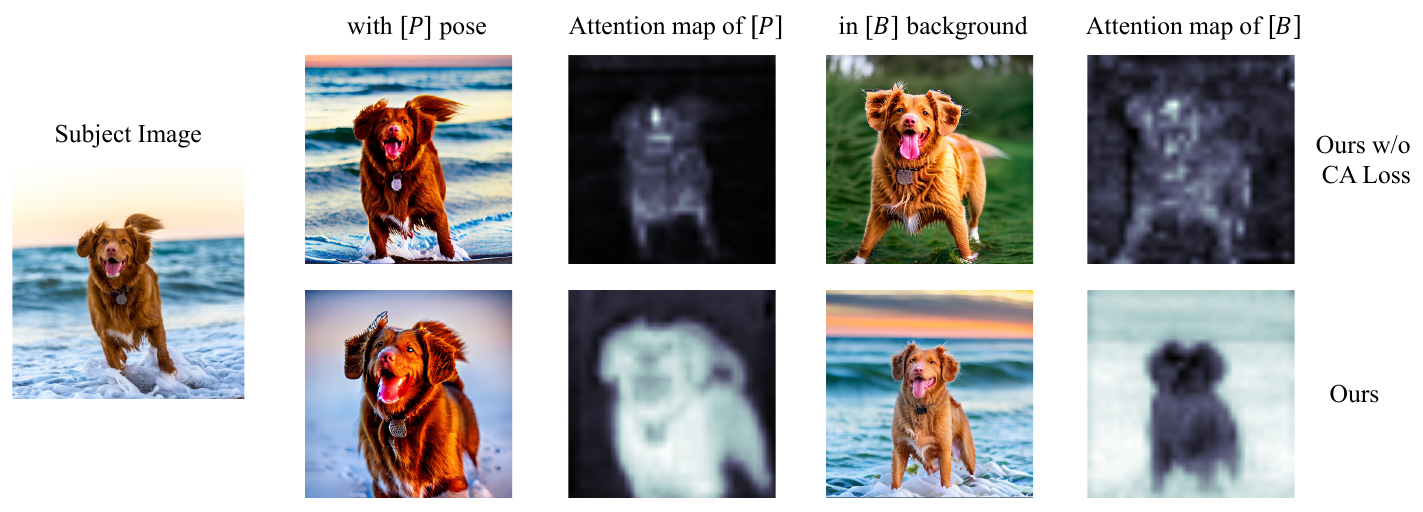}
    \vspace{-1.5em}
    \caption{\textbf{Visual comparisons on the effect of cross attention loss}. Without the cross-attention loss, the learned pose is entangled with background information.}
    \vspace{-0.5em}
    \label{fig:ablation_caloss}
\end{figure}

\begin{table}[t]
    \small
    % \captionsetup{font={small, stretch=1.1}, justification=raggedright}
    \centering
    \caption{\textbf{Ablation study.} With the proposed attribute mapper, attention loss $L_{CA}$, and joint training strategy, our DETEX achieves the best subject fidelity while maintaining a comparable text alignment performance.}
    \vspace{-0.5em}
    % \resizebox{\textwidth}{14mm}{
    \setlength{\tabcolsep}{1mm}{
    \label{tab4:ablation}  
    \begin{tabular}{cccc}
        \toprule
        Method & CLIP-T ($\uparrow$)& CLIP-I(FG) ($\uparrow$)
        & DINO-I(FG) ($\uparrow$)\\
        \midrule
        w/o Mapper & 0.3295 & 0.7563 & 0.5789 \\ 
        w/o Attn Loss & 0.3287 & 0.7548 & 0.5766 \\
        w/o Joint Training & \textbf{0.3327} & 0.7403 & 0.5625 \\
        Ours & 0.3301 & \textbf{0.7695}  &  \textbf{0.5797}\\
        \bottomrule
    \end{tabular}}
    \vspace{-1em}
\end{table}

\noindent \textbf{Effect of Attribute Mappers.}
We have conducted the ablation to evaluate the effect of the attribute mappers.
Specifically, we remove the mapper and optimize the pose and background embeddings directly.
As shown in Fig.~\ref{fig:ablation_mapper}, without the attribute mappers, the subject embedding tends to entangle with pose information, while the pose embedding entangles with background information.
In contrast, with the mappers, our method can disentangle each information successfully.
From Table~\ref{tab4:ablation}, our method with attribute mappers achieves better on both text and image alignment scores, demonstrating its effectiveness.

\noindent \textbf{Effect of Cross Attention Loss.}
We also study the effect of introduced cross-attention loss and compare it with the variant that is trained without $L_{CA}$.
As shown in Fig.~\ref{fig:ablation_caloss}, without the cross-attention loss, the subject-unrelated embeddings \texttt{[P]} and \texttt{[B]} fail to exhibit independent control over the pose and background, respectively. 
Neither \texttt{[P]} nor \texttt{[B]} has significant attention at the target region. 
After applying the attention constraint, our method performs better in the disentanglement of unrelated information, leading to higher text and image alignment scores (as shown in Table~\ref{tab4:ablation}).

\begin{figure}[t]
    \centering
    \includegraphics[width=1\linewidth]{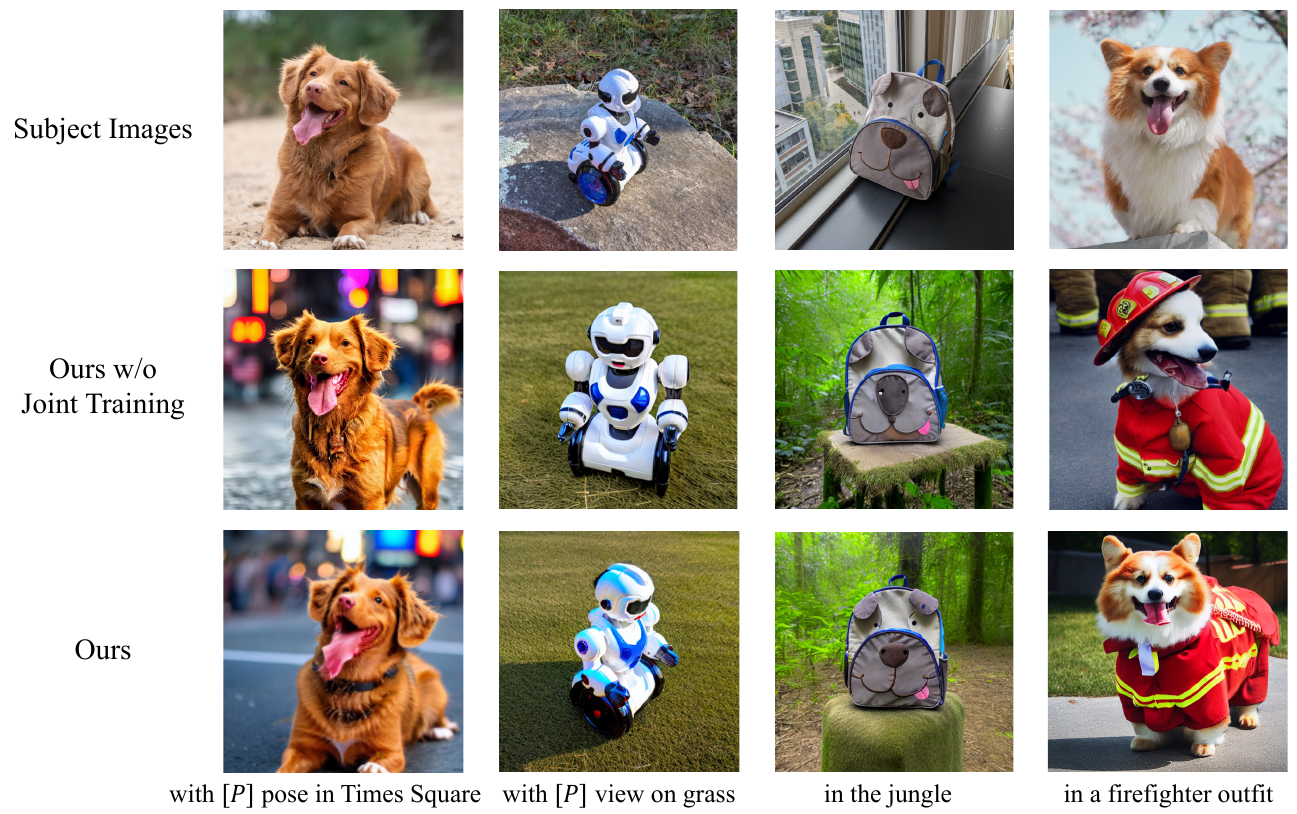}
    \vspace{-1.5em}
    \caption{\textbf{Visual comparisons on the effect of the joint training strategy}. Without the joint training, the learned concept tends to have inconsistent details with the given image (\textit{e.g}, the backpack in the $2rd$ row).}
    \vspace{-1em}
    \label{fig:ablation_joint}
\end{figure}

\noindent \textbf{Effect of the Joint Training Strategy.}
The joint training strategy effectively facilitates the decoupling between learned embeddings, resulting in better consistency between the learned subject embedding and the target concept. 
To demonstrate this, we additionally train our method without the joint training strategy (\textit{i.e}, $\gamma$ = 0).
As shown in Fig.~\ref{fig:ablation_joint}, without the joint training strategy, the learned pose and background embeddings are entangled, and the details of subject embedding (\texttt{[V]}) are also inconsistent with the target concept (\textit{e.g}, the backpack in $3rd$ column). 
From Table~\ref{tab4:ablation}, with the joint training strategy, better decoupling between the learned embeddings is achieved, resulting in improved maintenance of target concept consistency. 

% To clarify the effectiveness of the disentanglement between the unrelated embeddings, Fig.~\ref{fig:all} illustrates more editing results with various combinations of \texttt{[P]} and \texttt{[B]}. We also combine the unrelated tokens with non-target subjects, the results are shown in Fig.~\ref{fig_sup3}. The above results demonstrate the representation ability and sufficient decoupling of the unrelated tokens. 

% We have also studied the effect of the number of training images $N$. 
% %
% As the $N$ increases, the subject fidelity of learned concept improves, while slightly benefiting the editability.
% %
% More details and additional ablation studies can be found in the supplementary materials.

\section{5 Conclusion}

In this paper, we proposed a novel method, namely DETEX, for flexible customized text-to-image generation.
In particular, our DETEX adopts multiple decoupled textual embeddings to represent the subject information and the subject-unrelated pose and background information separately.
Our proposed joint training strategy and the cross-attention loss further facilitate the decoupling between the subject embedding and the irrelevant embeddings, while improving the consistency between the learned subject and the target concept.
Quantitative and qualitative evaluations demonstrate that our method outperforms the state-of-the-art methods in terms of both editing flexibility and subject fidelity. 
In future work, we
will try to investigate effective methods for the ability to adaptively identify meaningfully
attributes of the given subject (e.g., face attributes), thus facilitating a more flexible generation process.

\section{Acknowledgments}
This work was supported in part by National Key R\&D Program of China under Grant No. 2020AAA0104500, and the National Natural Science Foundation of China under Grant No.s 62176249 and U19A2073.
 
\bibliography{ref.bib}

\begin{thebibliography}{29}
\providecommand{\natexlab}[1]{#1}

\bibitem[{Avrahami et~al.(2023)Avrahami, Aberman, Fried, Cohen-Or, and Lischinski}]{avrahami2023break}
Avrahami, O.; Aberman, K.; Fried, O.; Cohen-Or, D.; and Lischinski, D. 2023.
\newblock Break-A-Scene: Extracting Multiple Concepts from a Single Image.
\newblock \emph{arXiv preprint arXiv:2305.16311}.

\bibitem[{Avrahami, Lischinski, and Fried(2022)}]{blended—diffusion}
Avrahami, O.; Lischinski, D.; and Fried, O. 2022.
\newblock Blended Diffusion for Text-driven Editing of Natural Images.
\newblock In \emph{2022 IEEE/CVF Conference on Computer Vision and Pattern Recognition (CVPR)}.

\bibitem[{Caron et~al.(2021)Caron, Touvron, Misra, Jegou, Mairal, Bojanowski, and Joulin}]{DINO}
Caron, M.; Touvron, H.; Misra, I.; Jegou, H.; Mairal, J.; Bojanowski, P.; and Joulin, A. 2021.
\newblock Emerging Properties in Self-Supervised Vision Transformers.
\newblock In \emph{2021 IEEE/CVF International Conference on Computer Vision (ICCV)}.

\bibitem[{Chen et~al.(2023)Chen, Zhang, Wang, Duan, Zhou, and Zhu}]{disenbooth}
Chen, H.; Zhang, Y.; Wang, X.; Duan, X.; Zhou, Y.; and Zhu, W. 2023.
\newblock DisenBooth: Disentangled Parameter-Efficient Tuning for Subject-Driven Text-to-Image Generation.
\newblock \emph{arXiv preprint arXiv:2305.03374}.

\bibitem[{Devlin et~al.(2018)Devlin, Chang, Lee, and Toutanova}]{BERT}
Devlin, J.; Chang, M.-W.; Lee, K.; and Toutanova, K. 2018.
\newblock Bert: Pre-training of deep bidirectional transformers for language understanding.
\newblock \emph{arXiv preprint arXiv:1810.04805}.

\bibitem[{Dhariwal and Nichol(2021)}]{diffusion}
Dhariwal, P.; and Nichol, A. 2021.
\newblock Diffusion models beat gans on image synthesis.
\newblock \emph{Advances in Neural Information Processing Systems}, 34: 8780--8794.

\bibitem[{Gal et~al.(2022)Gal, Alaluf, Atzmon, Patashnik, Bermano, Chechik, and Cohen-or}]{TI}
Gal, R.; Alaluf, Y.; Atzmon, Y.; Patashnik, O.; Bermano, A.~H.; Chechik, G.; and Cohen-or, D. 2022.
\newblock An Image is Worth One Word: Personalizing Text-to-Image Generation using Textual Inversion.
\newblock In \emph{The Eleventh International Conference on Learning Representations}.

\bibitem[{Han et~al.(2023)Han, Li, Zhang, Milanfar, Metaxas, and Yang}]{svdiff}
Han, L.; Li, Y.; Zhang, H.; Milanfar, P.; Metaxas, D.; and Yang, F. 2023.
\newblock Svdiff: Compact parameter space for diffusion fine-tuning.
\newblock \emph{arXiv preprint arXiv:2303.11305}.

\bibitem[{Hao et~al.(2023)Hao, Han, Zhao, and Wong}]{vico}
Hao, S.; Han, K.; Zhao, S.; and Wong, K.-Y.~K. 2023.
\newblock ViCo: Detail-Preserving Visual Condition for Personalized Text-to-Image Generation.
\newblock \emph{arXiv preprint arXiv:2306.00971}.

\bibitem[{Ho, Jain, and Abbeel(2020)}]{DDPM}
Ho, J.; Jain, A.; and Abbeel, P. 2020.
\newblock Denoising Diffusion Probabilistic Models.
\newblock \emph{Neural Information Processing Systems}.

\bibitem[{Ho and Salimans(2022)}]{classifier-free}
Ho, J.; and Salimans, T. 2022.
\newblock Classifier-free diffusion guidance.
\newblock \emph{arXiv preprint arXiv:2207.12598}.

\bibitem[{Hu et~al.(2021)Hu, Wallis, Allen-Zhu, Li, Wang, Wang, Chen et~al.}]{lora}
Hu, E.~J.; Wallis, P.; Allen-Zhu, Z.; Li, Y.; Wang, S.; Wang, L.; Chen, W.; et~al. 2021.
\newblock LoRA: Low-Rank Adaptation of Large Language Models.
\newblock In \emph{International Conference on Learning Representations}.

\bibitem[{Kim, Kwon, and Ye(2022)}]{DiffusionCLIP}
Kim, G.; Kwon, T.; and Ye, J.~C. 2022.
\newblock Diffusionclip: Text-guided diffusion models for robust image manipulation.
\newblock In \emph{Proceedings of the IEEE/CVF Conference on Computer Vision and Pattern Recognition}, 2426--2435.

\bibitem[{Kirillov et~al.(2023)Kirillov, Mintun, Ravi, Mao, Rolland, Gustafson, Xiao, Whitehead, Berg, Lo et~al.}]{SAM}
Kirillov, A.; Mintun, E.; Ravi, N.; Mao, H.; Rolland, C.; Gustafson, L.; Xiao, T.; Whitehead, S.; Berg, A.~C.; Lo, W.-Y.; et~al. 2023.
\newblock Segment anything.
\newblock \emph{arXiv preprint arXiv:2304.02643}.

\bibitem[{Kumari et~al.(2023)Kumari, Zhang, Zhang, Shechtman, and Zhu}]{CD}
Kumari, N.; Zhang, B.; Zhang, R.; Shechtman, E.; and Zhu, J.-Y. 2023.
\newblock Multi-concept customization of text-to-image diffusion.
\newblock In \emph{Proceedings of the IEEE/CVF Conference on Computer Vision and Pattern Recognition}, 1931--1941.

\bibitem[{Liu et~al.(2023{\natexlab{a}})Liu, Park, Azadi, Zhang, Chopikyan, Hu, Shi, Rohrbach, and Darrell}]{classfier-guaidance}
Liu, X.; Park, D.~H.; Azadi, S.; Zhang, G.; Chopikyan, A.; Hu, Y.; Shi, H.; Rohrbach, A.; and Darrell, T. 2023{\natexlab{a}}.
\newblock More Control for Free! Image Synthesis with Semantic Diffusion Guidance.
\newblock In \emph{2023 IEEE/CVF Winter Conference on Applications of Computer Vision (WACV)}.

\bibitem[{Liu et~al.(2023{\natexlab{b}})Liu, Feng, Zhu, Zhang, Zheng, Liu, Zhao, Zhou, and Cao}]{cones}
Liu, Z.; Feng, R.; Zhu, K.; Zhang, Y.; Zheng, K.; Liu, Y.; Zhao, D.; Zhou, J.; and Cao, Y. 2023{\natexlab{b}}.
\newblock Cones: Concept neurons in diffusion models for customized generation.
\newblock \emph{arXiv preprint arXiv:2303.05125}.

\bibitem[{Loshchilov and Hutter(2018)}]{adamw}
Loshchilov, I.; and Hutter, F. 2018.
\newblock Decoupled Weight Decay Regularization.
\newblock In \emph{International Conference on Learning Representations}.

\bibitem[{Nichol et~al.(2022)Nichol, Dhariwal, Ramesh, Shyam, Mishkin, Mcgrew, Sutskever, and Chen}]{GLIDE}
Nichol, A.~Q.; Dhariwal, P.; Ramesh, A.; Shyam, P.; Mishkin, P.; Mcgrew, B.; Sutskever, I.; and Chen, M. 2022.
\newblock GLIDE: Towards Photorealistic Image Generation and Editing with Text-Guided Diffusion Models.
\newblock In \emph{International Conference on Machine Learning}, 16784--16804.

\bibitem[{Radford et~al.(2021)Radford, Kim, Hallacy, Ramesh, Goh, Agarwal, Sastry, Askell, Mishkin, Clark et~al.}]{CLIP}
Radford, A.; Kim, J.~W.; Hallacy, C.; Ramesh, A.; Goh, G.; Agarwal, S.; Sastry, G.; Askell, A.; Mishkin, P.; Clark, J.; et~al. 2021.
\newblock Learning transferable visual models from natural language supervision.
\newblock In \emph{International conference on machine learning}, 8748--8763.

\bibitem[{Ramesh et~al.(2022)Ramesh, Dhariwal, Nichol, Chu, and Chen}]{DALLE2}
Ramesh, A.; Dhariwal, P.; Nichol, A.; Chu, C.; and Chen, M. 2022.
\newblock Hierarchical text-conditional image generation with clip latents.
\newblock \emph{arXiv preprint arXiv:2204.06125}.

\bibitem[{Rombach et~al.(2022)Rombach, Blattmann, Lorenz, Esser, and Ommer}]{LDM}
Rombach, R.; Blattmann, A.; Lorenz, D.; Esser, P.; and Ommer, B. 2022.
\newblock High-resolution image synthesis with latent diffusion models.
\newblock In \emph{Proceedings of the IEEE/CVF Conference on Computer Vision and Pattern Recognition}, 10684--10695.

\bibitem[{Ruiz et~al.(2023)Ruiz, Li, Jampani, Pritch, Rubinstein, and Aberman}]{dreambooth}
Ruiz, N.; Li, Y.; Jampani, V.; Pritch, Y.; Rubinstein, M.; and Aberman, K. 2023.
\newblock Dreambooth: Fine tuning text-to-image diffusion models for subject-driven generation.
\newblock In \emph{Proceedings of the IEEE/CVF Conference on Computer Vision and Pattern Recognition}, 22500--22510.

\bibitem[{Saharia et~al.(2022)Saharia, Chan, Saxena, Li, Whang, Denton, Ghasemipour, Gontijo~Lopes, Karagol~Ayan, Salimans et~al.}]{Imagen}
Saharia, C.; Chan, W.; Saxena, S.; Li, L.; Whang, J.; Denton, E.~L.; Ghasemipour, K.; Gontijo~Lopes, R.; Karagol~Ayan, B.; Salimans, T.; et~al. 2022.
\newblock Photorealistic text-to-image diffusion models with deep language understanding.
\newblock \emph{Advances in Neural Information Processing Systems}, 35: 36479--36494.

\bibitem[{Schuhmann et~al.(2021)Schuhmann, Vencu, Beaumont, Kaczmarczyk, Mullis, Katta, Coombes, Jitsev, and Komatsuzaki}]{LAION-400M}
Schuhmann, C.; Vencu, R.; Beaumont, R.; Kaczmarczyk, R.; Mullis, C.; Katta, A.; Coombes, T.; Jitsev, J.; and Komatsuzaki, A. 2021.
\newblock Laion-400m: Open dataset of clip-filtered 400 million image-text pairs.
\newblock \emph{arXiv preprint arXiv:2111.02114}.

\bibitem[{Shi et~al.(2023)Shi, Xiong, Lin, and Jung}]{Instantbooth}
Shi, J.; Xiong, W.; Lin, Z.; and Jung, H.~J. 2023.
\newblock Instantbooth: Personalized text-to-image generation without test-time finetuning.
\newblock \emph{arXiv preprint arXiv:2304.03411}.

\bibitem[{Song, Meng, and Ermon(2020)}]{DDIM}
Song, J.; Meng, C.; and Ermon, S. 2020.
\newblock Denoising Diffusion Implicit Models.
\newblock In \emph{International Conference on Learning Representations}.

\bibitem[{Vaswani et~al.(2017)Vaswani, Shazeer, Parmar, Uszkoreit, Jones, Gomez, Kaiser, and Polosukhin}]{attention}
Vaswani, A.; Shazeer, N.; Parmar, N.; Uszkoreit, J.; Jones, L.; Gomez, A.~N.; Kaiser, {\L}.; and Polosukhin, I. 2017.
\newblock Attention is all you need.
\newblock \emph{Advances in neural information processing systems}, 30.

\bibitem[{Wei et~al.(2023)Wei, Zhang, Ji, Bai, Zhang, and Zuo}]{ELITE}
Wei, Y.; Zhang, Y.; Ji, Z.; Bai, J.; Zhang, L.; and Zuo, W. 2023.
\newblock ELITE: Encoding Visual Concepts into Textual Embeddings for Customized Text-to-Image Generation.

\end{thebibliography}

\clearpage
\appendix
\section{Supplementary Material}
\section{A. More Experimental Details}

\subsection{A.1 Training Details}

\subsubsection{Our DETEX.}

We train our method and the baselines with Stable Diffusion v1.4. 
The regularization datasets are generated through vanilla Stable Diffusion using a class word (\textit{e.g}, ``\texttt{a dog}'') with 50 DDIM~\cite{DDIM} steps.
For each category, we generate 200 images as the regularization datasets.
During the evaluation of the CLIP-T metric, the learnable token in the generation prompts were removed. 
For methods \cite{dreambooth, CD, svdiff} that treat the learnable token as an identifier, the generation prompt like ``\texttt{a [V] dog on the beach}" would be transformed into ``\texttt{a dog on the beach}". 
For methods without category words in the training prompts, such as \cite{TI,vico}, we replace the learnable tokens with the corresponding category word. 
For example, the generation prompt ``\texttt{a * on the beach}" would be transformed into ``\texttt{a dog on the beach"}. 

% The CLIP-T metric was then calculated using the transformed prompts.

% Our method and the baselines are all based on Stable Diffusion v1.4 implementations of the HuggingFace diffusers library. The regularization datasets are generated through vanilla Stable Diffusion using a class word (\textit{e.g}, ``\texttt{a dog}'') with 50 DDIM~\cite{DDIM} steps, consisting of 200 images for each category. During the evaluation of the CLIP-T metric, the learnable token in the generation prompts were removed. For methods \cite{dreambooth, CD, svdiff} that treat the learnable token as an identifier, generation prompt like ``\texttt{a [V] dog on the beach}" would be transformed into ``\texttt{a dog on the beach}". For methods \cite{TI,vico} that didn't employ category words in the training prompts, we replaced the learnable tokens with the corresponding category word. For example, the generation prompt ``\texttt{a * on the beach}" would be transformed into ``\texttt{a dog on the beach"}. The CLIP-T metric was then calculated using the transformed prompts.

\subsubsection{Textual Inversion~\cite{TI}.}

We utilize the official open-source code of Textual Inversion\footnote{https://github.com/rinongal/textual\_inversion}.
We train it for 5000 steps with a learning rate of 0.005 and a batch size of 1. 
The category word for each subject is employed to initialize the learnable token. 
In cases where a category was represented by multiple tokens (\textit{e.g}, ``\texttt{bear plushie}“), we use a single token that captured the similar meaning (\textit{e.g}, ``\texttt{plushie}”). 
For training text prompts, we adopt the single token templates set provided by the official code.

% We utilized the official open-source code, training for 5000 steps with a learning rate of 0.005 and a batch size of 1. We employed the category word for each subject to initialize the learnable token. In cases where a category was represented by multiple tokens (\textit{e.g}, ``\texttt{bear plushie}“), we used a single token that captured the similar meaning (\textit{e.g}, ``\texttt{plushie}”). For training text prompts, we adopt the single token templates set provided by the official code.

\subsubsection{Dreambooth~\cite{dreambooth}.}

As the official code for Dreambooth is not publicly accessible, we employ a third-party implementation\footnote{https://github.com/XavierXiao/Dreambooth-Stable-Diffusion}. 
The text transformer is frozen and the parameters of the U-Net model are fine-tuned during training. 
We set the learning rate to 1e-6 and train the model for 1000 steps with a batch size of 1.

% As the official code for Dreambooth is not publicly accessible, we employ a third-party implementation provided by XavierXiao on GitHub. The text transformer is frozen and the parameters of the U-Net model are fine-tuned during training. We set the learning rate to 1e-6 and train the model for 1000 steps with a batch size of 1.

\subsubsection{Custom Diffusion~\cite{CD}.}

We employ the official open-source code of Custom Diffusion\footnote{https://github.com/adobe-research/custom-diffusion}.
We use the default hyperparameter settings with a learning rate of 1e-5 and a batch size of 8 for training over 300 steps. 
During training, only the K, V layer within the cross-attention module of the U-Net and the learnable embeddings are fine-tuned. 
%
% The training prompt employed is ``\texttt{Photo of a [V] category}" and for the data augmentation of zooming in or out, captions like ``\texttt{a far away}" or ``\texttt{a very small}" were added to the beginning of the prompt.

% We adopted the hyperparameter settings from the official code, with a learning rate of 1e-5 and a batch size of 8 for training over 300 steps. During training, only the K, V layer within the cross-attention module of the U-Net and the learnable tokens were fine-tuned. The training prompt employed was ``\texttt{Photo of a [V] category}" and for the data augmentation of zooming in or out, captions like ``\texttt{a far away}" or ``\texttt{a very small}" were added to the beginning of the prompt.

\subsubsection{ViCo~\cite{vico}.}

We employed the official open-source code of ViCo\footnote{https://github.com/haoosz/ViCo}, adhering to the default hyperparameter settings. 
Specifically, the learning rate and batch size are set to 0.005 and 4, and training is carried out for 400 steps.
The training prompt employed is ``\texttt{Photo of a *}", as ViCo treats the target concept as a single token to achieving effective text-image alignment. 
During the inference, we utilize the checkpoint trained at 300 steps, and a random input image was selected as the subject image.

% We employed the official open-source code, adhering to the default hyperparameter settings. Specifically, the learning rate was set to 0.005, batch size to 4, and training was carried out for 400 steps. The training prompt employed was ``\texttt{Photo of a *}", as ViCo treats the target concept as a single token to achieving effective text-image alignment. During the inference, we utilized the checkpoint trained at 300 steps, and a random input image was selected as the reference image.

\subsubsection{SVDiff~\cite{svdiff}.}

We utilize the official open-source code of SVDiff~\footnote{https://github.com/mkshing/svdiff-pytorch} and fine-tune both the spectral shift and 1-D weight kernels. 
The learning rate for spectral shifts is set to 1e-3, and for 1-D weight kernels, it is set to 1e-6. 
The training is conducted with a batch size of 2 for 1000 steps. 
The training prompt employed is ``\texttt{Photo of a [V] class”}.

% We utilize the official open-source code and fine-tuned both the spectral shift and 1-D weight kernels. The learning rate for spectral shifts was set to 1e-3, and for 1-D weight kernels, it was set to 1e-6. The training was conducted with a batch size of 2 for 1000 steps. The training
% prompt employed was ``\texttt{Photo of a [V] class”}.

\begin{figure*}[htp]
    \centering
    \includegraphics[width=1\linewidth]{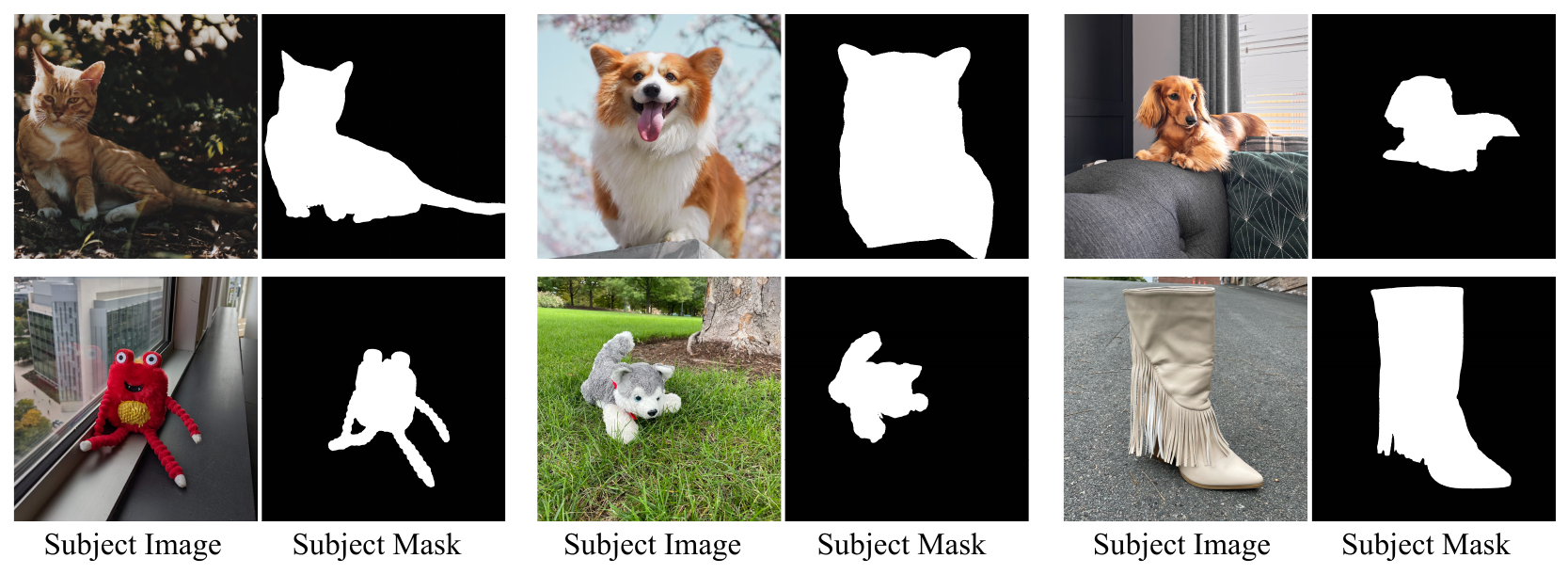}
    \caption{\textbf{Subject masks used in our method}. For each pair, on the left is the input image, and on the right is the corresponding mask obtained through pre-trained SAM.}
    \label{fig_mask}
\end{figure*}

\subsection{A.2 Subject Masks}

We adopt the pre-trained segmentation model SAM~\cite{SAM} to simply obtain the subject masks of given images. 
Specifically, we utilize the pre-trained SAM to obtain initial foreground masks and then manually selected the target concept masks. 
Fig.~\ref{fig_mask} illustrates the training examples used in our experiments.

\begin{table}[htbp]
    \scriptsize
    \captionsetup{font={stretch=1.25}, justification=raggedright}
    \centering
    \caption{Quantitative comparisons with model trained with masked diffusion loss.}
    % \resizebox{\textwidth}{14mm}{
    \setlength{\tabcolsep}{1mm}{
    \label{tab:mask_loss}  
    \begin{tabular}{cccc}
        \toprule
        Method & CLIP-T ($\uparrow$)& CLIP-I(FG) ($\uparrow$)
        & DINO-I(FG) ($\uparrow$)\\
        \midrule
        Ours & \textbf{0.3301} & \textbf{0.7695}  &  \textbf{0.5797}\\
        + Masked Diffusion Loss & 0.3289 & 0.7498 & 0.5624 \\
        \bottomrule
    \end{tabular}}
\end{table}

\begin{figure*}[t]
    \centering
    \includegraphics[width=1\linewidth]{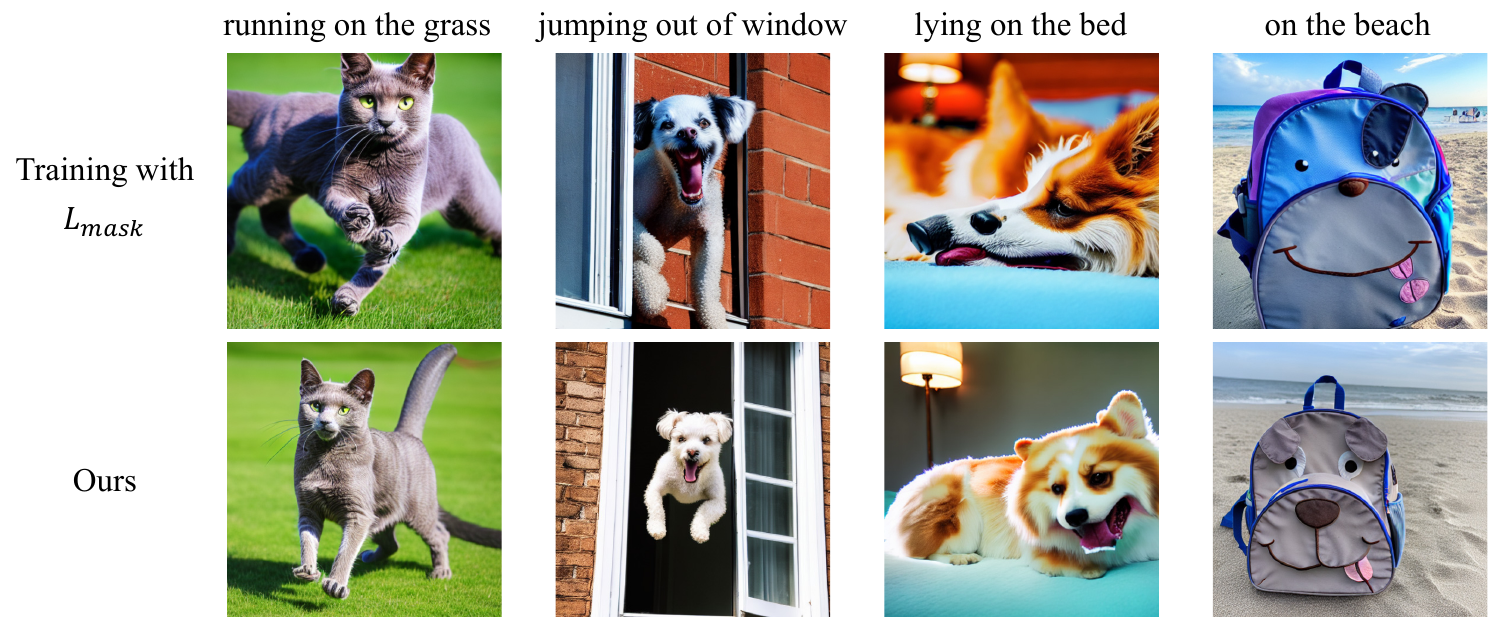}
    \caption{\textbf{Visualization results on the effect of masked diffusion loss}. After applying the masked diffusion loss, obvious artifacts appear in the editing of poses (see columns 1 and 3), while the texture or colour of the target concept is hard to maintain (see columns 2 and 4).}
    \label{fig_mask_loss}
\end{figure*}

\begin{figure*}[ht]
    \centering
    \includegraphics[width=1\linewidth]{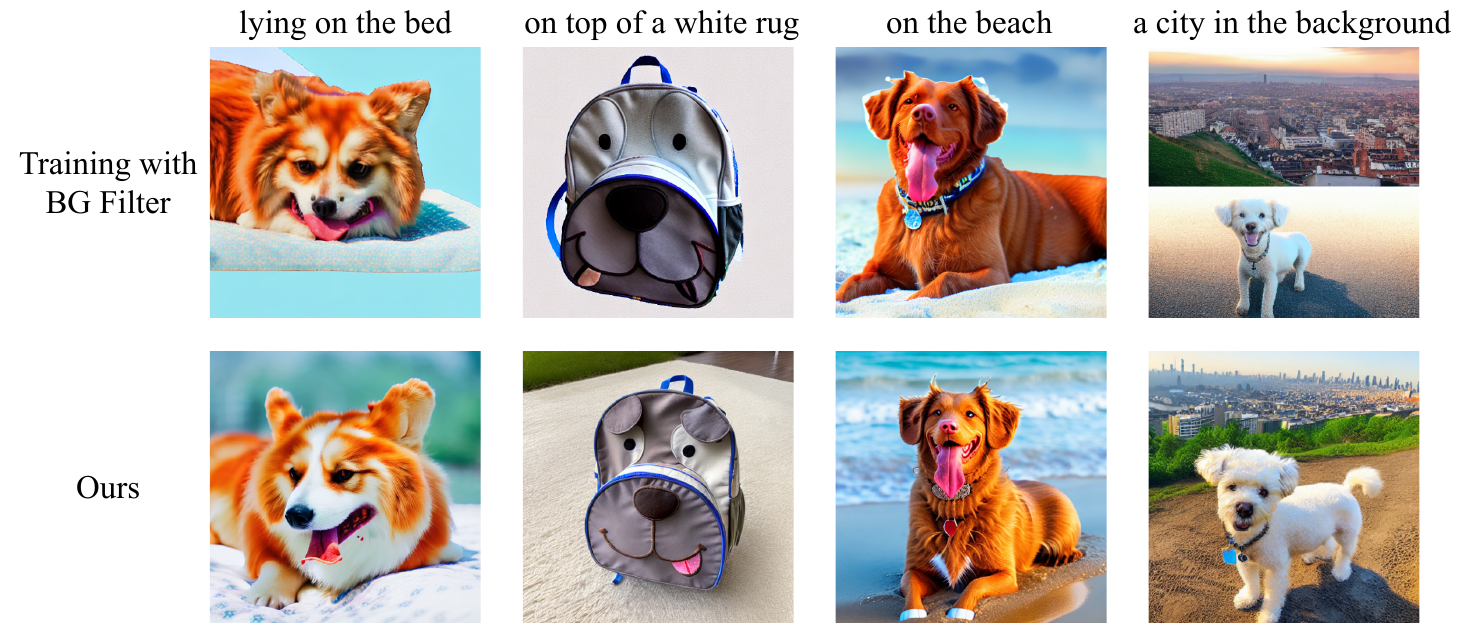}
    \caption{\textbf{Results compared with background filtration.} Completely discarding the background during training can lead the model to overfit on blank backgrounds, which in turn introduces a significant amount of unnatural solid-color backgrounds in the generated results.}
    \label{fig_only_mask}
\end{figure*}

\begin{figure*}[t]
    \centering
    \includegraphics[width=1\linewidth]{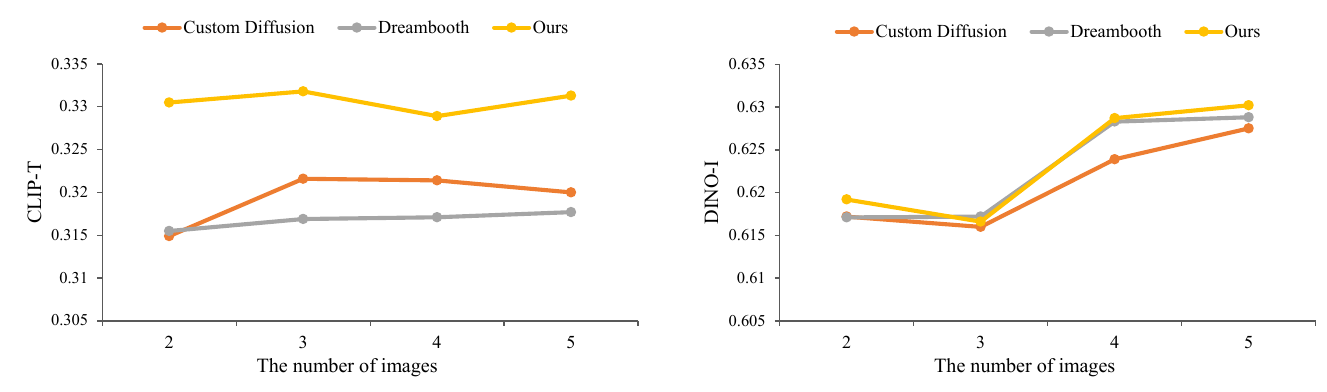}
    \caption{\textbf{Left:} Text fidelity compared with the effect of input image number on CLIP-T. \textbf{Right:} Text fidelity compared with the effect of input image number on DINO-I (FG)}
    \label{fig_number}
\end{figure*}

\section{B. Additional Experiments}

\subsection{B.1 More Ablation Study}

\subsubsection{Comparison with Masked Diffusion Loss.}

\cite{CD, avrahami2023break} have employed the technique that the diffusion loss is calculated only within the spatial area of the target region, disregarding the influence of content in other regions during the optimization process:
\begin{equation}
\small
L_{mask} = \mathbb{E}_{z\sim\mathcal{E}(x), y, \epsilon \sim \mathcal{N}(0, 1), t }\Big[ \Vert \left[ \epsilon - \epsilon_\theta(z_{t},t, \tau_\theta(y) \right] \odot M) \Vert_{2}^{2}\Big] \, ,
\label{eq:masked_loss}
\end{equation}
where $M$ denotes the mask of the target region.
We have conducted the experiment by applying this masked diffusion loss to background-filtered training steps using a foreground mask, specifically during the reconstruction of foreground-only images. 
However, based on experimental results, we found that this technique would reduce the accuracy of \texttt{[V]} in reconstructing the target concept and also disrupt the pose of living entities in the generated results. 
As the visualization results are shown in Fig.~\ref{fig_mask_loss}, the living subjects exhibited unrealistic poses, while the textures and details of subjects were compromised and cannot be consistent with the inputs. 
The quantitative evaluation metrics of using this technique (shown in Table~\ref{tab:mask_loss}) are also declined compared to our method. 

% Consequently, we decided not to employ the masked diffusion loss in our DETEX.

% We have noticed a technique employed in \cite{CD, avrahami2023break}, where the diffusion loss is calculated only within the spatial area of the target region, disregarding the influence of content in other regions during the optimization process:
% \begin{equation}
% \small
% L_{mask} = \mathbb{E}_{z\sim\mathcal{E}(x), y, \epsilon \sim \mathcal{N}(0, 1), t }\Big[ \Vert [ \epsilon M - \epsilon_\theta(z_{t},t, \tau_\theta(y) ] \odot M) \Vert_{2}^{2}\Big] \, ,
% \label{eq:masked_loss}
% \end{equation}
% where $M$ denotes the mask of the target region.
% We tried to apply this masked diffusion loss to background-filtered training steps using a foreground mask, specifically during the reconstruction of foreground-only images. However, based on experimental results, we found that this technique would reduce the accuracy of \texttt{[V]} in reconstructing the target concept and also disrupt the pose of living entities in the generated results. As the visualization results shown in Fig.~\ref{fig_mask_loss}, the living subjects exhibited unrealistic poses, while textures and details of subjects were compromised and cannot be consistent with the inputs. The quantitative evaluation metrics of using this technique (shown in Tab.~\ref{tab:mask_loss}) are also declined compared to our method. Consequently, we decided not to employ the masked diffusion loss in our DETEX.

\begin{figure*}[htp]
    \centering
    \includegraphics[width=1\linewidth]{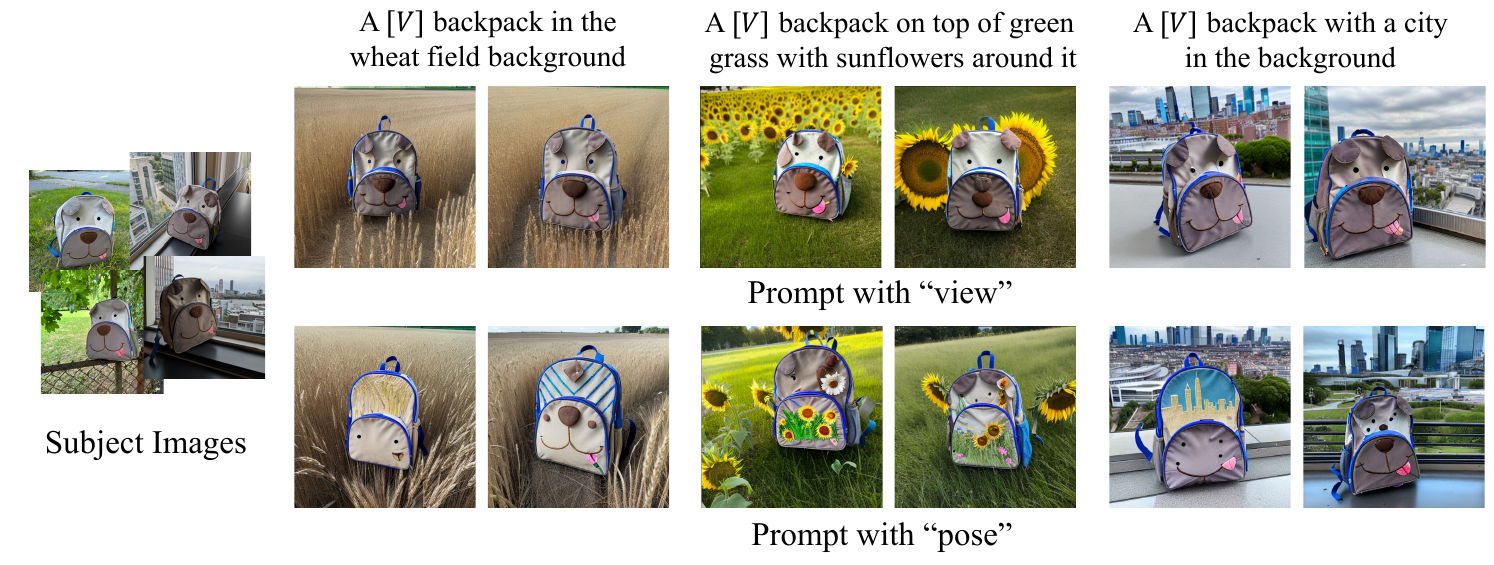}
    \caption{The visualization results on the different attribute words. For each row, we present the generation results with different attribute words used in the training prompt.}
    \label{fig_sup4}
\end{figure*}
\subsubsection{Comparison with Background Filtration.}

Previous methods~\cite{ELITE,avrahami2023break} considered employing a subject mask to completely filter out background information in order to avoid overfitting to irrelevant information. 
However, we found that this approach can lead the model to overfit to blank backgrounds, introducing artifacts in the generated images. 
We have conducted experiments by completely filtering out the background in the training process of DETEX (\textit{i.e}, set $\gamma$ = 1.0), and the resulting comparison of the generated effects is shown in Fig.~\ref{fig_only_mask}. 
An excessive amount of blank background can also lead to the generation of solid-color backgrounds and noticeable outlines of foreground objects (see the first row). 
Our joint training strategy alternates between presenting images with blank backgrounds and original backgrounds to the model. 
This approach effectively mitigates overfitting to blank backgrounds and facilitates the decoupling of background information.

% Previous methods \cite{ELITE,avrahami2023break} considered employing a subject mask to completely filter out background information in order to avoid overfitting to irrelevant information. However, we found that this approach can lead the model to overfit to blank backgrounds, introducing artifacts in the generated images. We tried to completely filter out the background in the training process of DETEX (\textit{i.e}, set $\gamma$ = 1.0), and the resulting comparison of the generated effects is shown in Fig. \ref{fig_only_mask}. An excessive amount of blank background can also lead to the generation of solid-color backgrounds and noticeable outlines of foreground objects (see the first row). Our joint training strategy alternates between presenting images with blank backgrounds and original backgrounds to the model. This approach effectively mitigates overfitting to blank backgrounds and facilitates the decoupling of background information.

\subsubsection{Effect of the Attribute Word.}

We observed that the attribute word for unrelated pseudo-word in training prompts can sufficiently guide the separation between the target concept and the irrelevant attributes. 
Fig.~\ref {fig_sup4} shows an example using a static subject, a backpack, training with different attribute words. 
When ``\texttt{view"} is used as the attribute word to assist the learning of \texttt{[P]}, the shared token \texttt{[V]} can fully capture the features of the target concept and maintains the consistency during editing (see the $1st$ column). 
However, when ``\texttt{pose"} is used as the attribute word to assist \texttt{[P]}'s learning, \texttt{[P]} overfits to some features of the backpack itself, thus affecting \texttt{[V]}'s representation of the target concept. 
Specifically, when ``\texttt{pose"} is used, the texture of the subject is prone to change. 
One possible explanation is that the pose guiding \texttt{[P]} to capture the facial expression on the backpack, \textit{i.e}, the pattern of a smiling dog. 
This phenomenon highlights the significant guiding effect of prompts during the training. 
In our proposed method, the design of prompts for training the tokens is rational and effective, as it allows \texttt{[V]} to learn the target concept accurately and guide the unrelated tokens to capture the corresponding irrelevant information effectively.

\subsubsection{Effect of the Number of Input Images.}

\begin{figure*}[htp]
    \centering
    \includegraphics[width=1\linewidth]{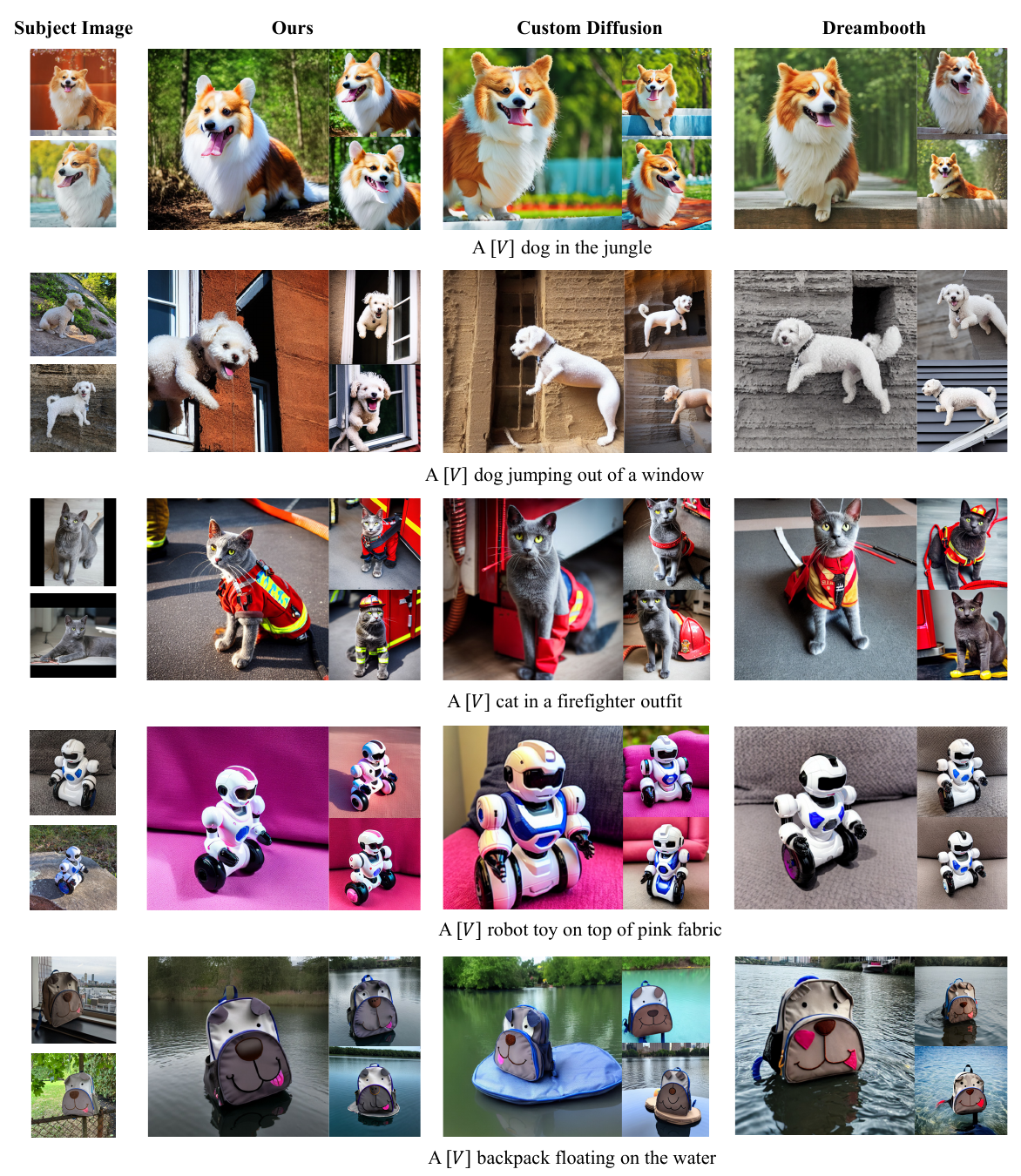}
    \caption{Qualitative comparisons of different methods with only 2 input images.}
    \label{fig_sup5}
\end{figure*}

We have investigated the effect of the number of input images on the model's performance. 
We select a subset on Dreambench with 5 subjects. 
For each subject, we randomly selected 2 to 5 images as inputs, and then compared the performance of several methods under different numbers of input images. 
Fig.~\ref{fig_number} shows the quantitative results regarding the effect of the number of input images on text fidelity and subject fidelity, respectively. 
Our method exhibits remarkable text fidelity regardless of the number of input images, demonstrating strong editing flexibility even in one-shot learning scenarios. 
However, subject fidelity appears to be more sensitive to the number of input images. 
In one-shot scenarios, maintaining subject consistency becomes challenging, as our method struggles to distinguish between shared concepts and image-specific concepts, leading to entanglement between the word \texttt{[V]}, \texttt{[P]} and \texttt{[B]}. 
However, when the number of input images exceeds one, the distinction between shared and image-specific concepts becomes evident, enabling the words to effectively capture the corresponding information and achieve better subject fidelity.
We also show the generation results with only two input images in Fig.~\ref{fig_sup5}. 
Comparatively, our method strikes a balanced performance between text fidelity and object fidelity with only two input images, showcasing great potential when dealing with extremely limited input images. 
This ability underscores the robustness and versatility of our method, making it a promising approach for personalized image generation tasks in various practical scenarios.

\subsection{B.2 More Results on Multi-Words Representation}
\begin{figure*}[htp]
    \centering
    \includegraphics[width=1\linewidth]{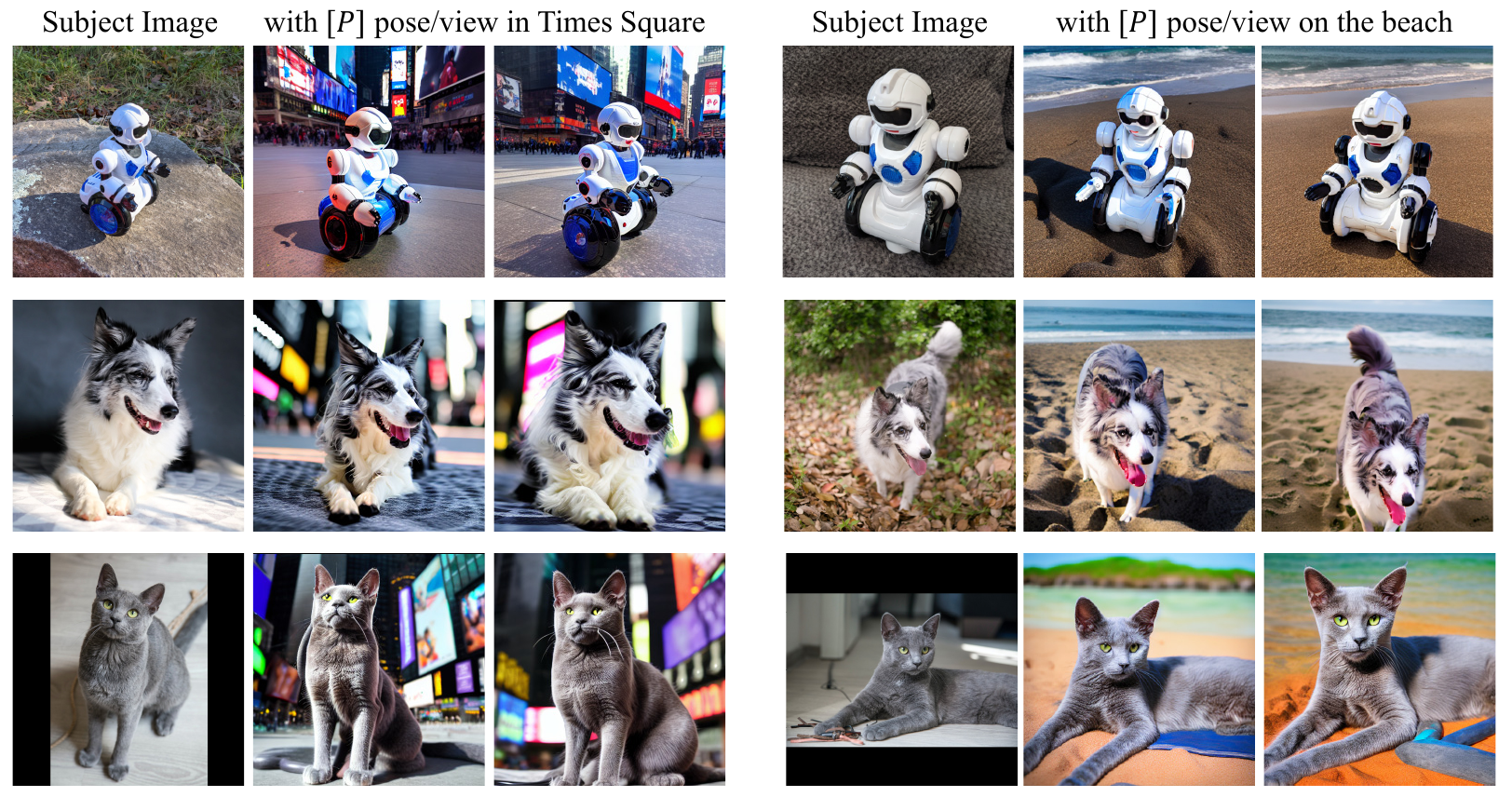}
    \caption{The generation results with pose (view) preservation.}
    \label{fig_sup1}
\end{figure*}

\begin{figure*}[htp]
    \centering
    \includegraphics[width=1\linewidth]{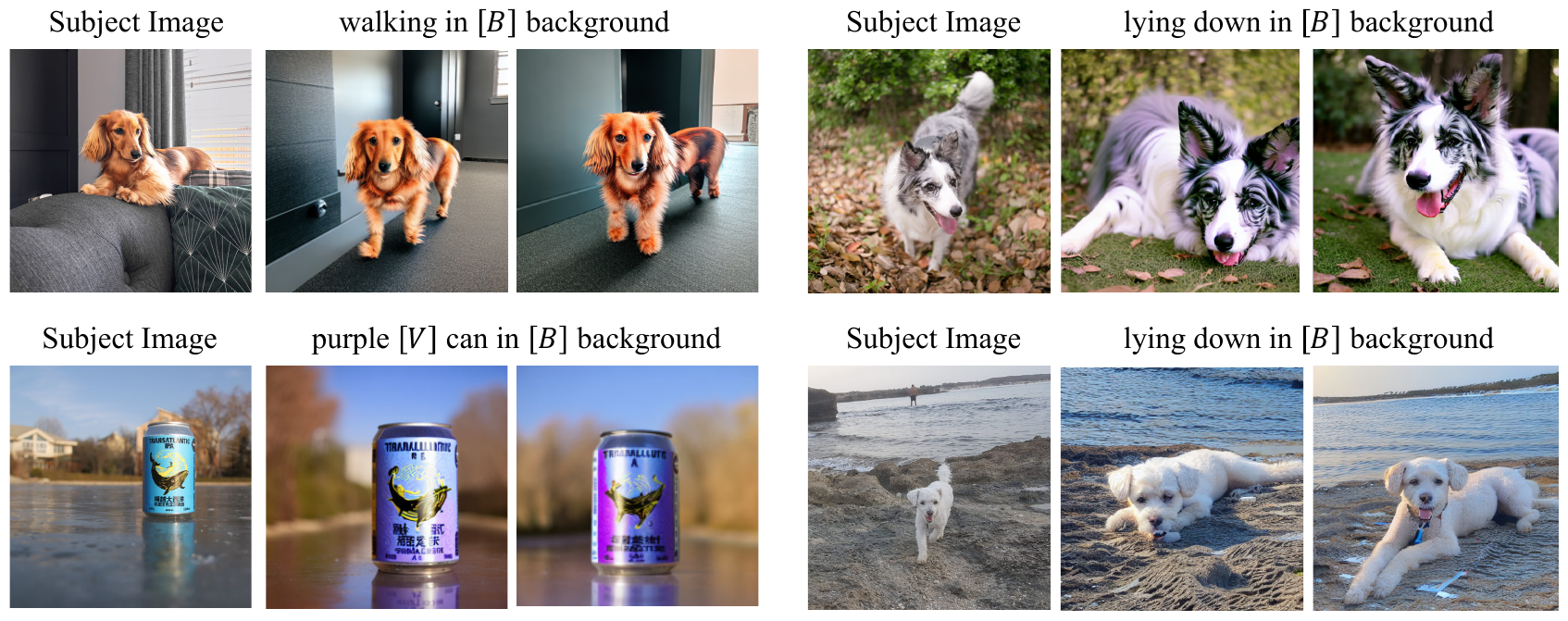}
    \caption{The generation results with background preservation.}
    \label{fig_sup2}
\end{figure*}

\begin{figure*}[htb]
    \centering
    \includegraphics[width=1\linewidth]{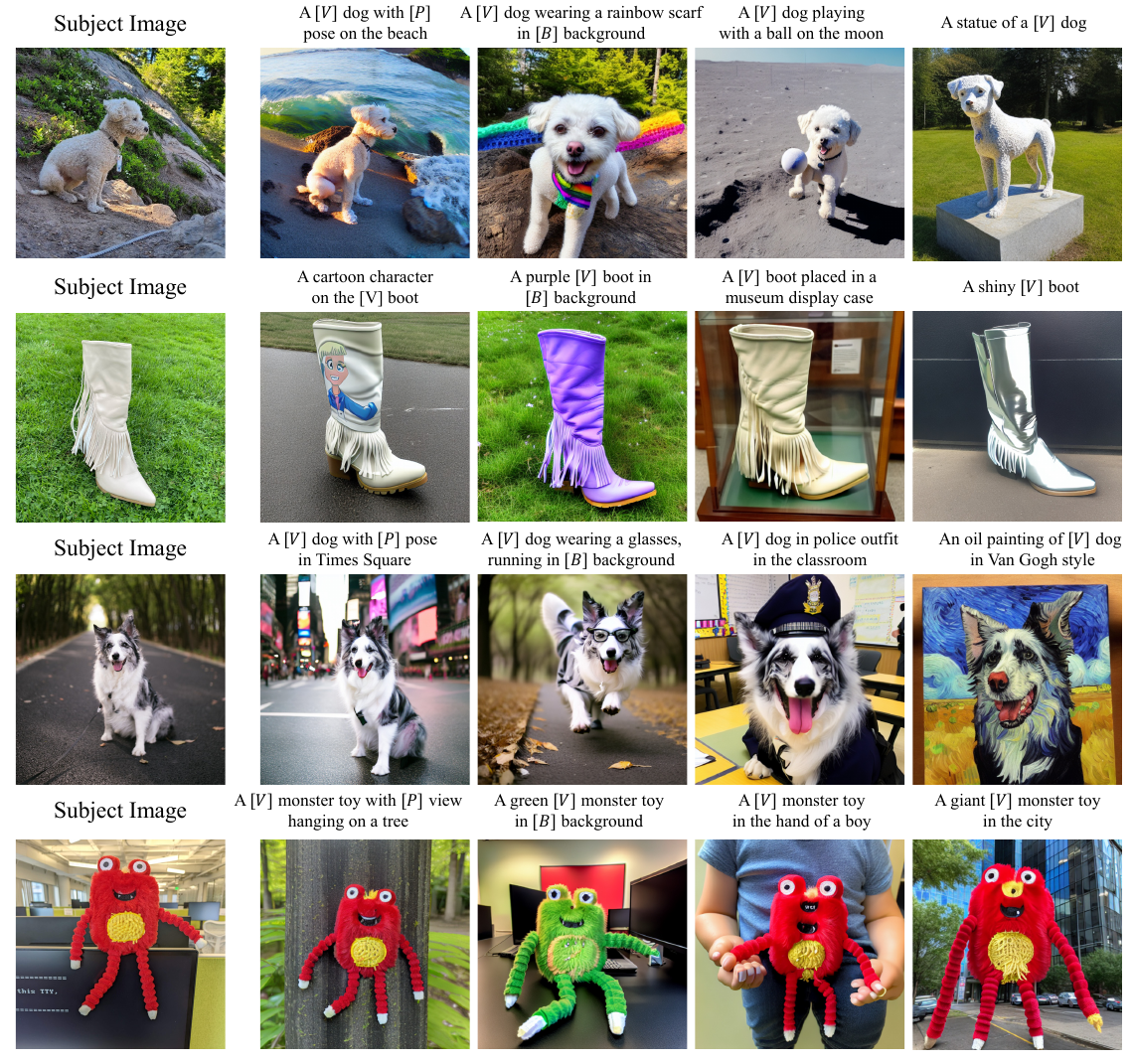}
    \caption{We perform more complex generations for the reference image shown in the first column. The $[P]$ and $[B]$ represent the corresponding pose (view) and background of the reference. The above results are trained on 4 input images.}
    \label{fig_app}
\end{figure*}

In this section, we show additional visualization editing results using irrelevant tokens. 
Fig.~\ref{fig_sup1} shows the editing results with pose (view) preservation. The pose (view) should be the same as the reference image, while the background should be changed. 
Fig.~\ref{fig_sup2} shows the editing results with background preservation, which is opposite to the previous one. 
In Fig.~\ref{fig_app}, we show more generation results with complex editing prompts.

% \bibliography{ref.bib}

\end{document}